\documentclass[letterpaper]{article} 
\usepackage{aaai2026}  
\usepackage{times}  
\usepackage{helvet}  
\usepackage{courier}  
\usepackage[hyphens]{url}  
\usepackage{graphicx} 
\usepackage{natbib}  
\usepackage{caption} 
\usepackage{algorithm}
\usepackage{algorithmic}

\usepackage{subfig}
\usepackage{amsmath, amssymb, amsthm}
\newtheorem{statement}{Statement}
\usepackage{tikz}
\usepackage{adjustbox}
\usepackage{multirow}
\usepackage{tabularx} 
\newcolumntype{C}{>{\centering\arraybackslash}X} 
\usepackage{xcolor}
\usepackage{booktabs}
\usepackage{array}
\usepackage{nameref}

\newcommand{\labelname}[1]{
  \def\@currentlabelname{#1}}%
\usepackage{subcaption}
\usepackage{newfloat}
\usepackage{listings}
\DeclareCaptionStyle{ruled}{labelfont=normalfont,labelsep=colon,strut=off} 
\floatstyle{ruled}
\newfloat{listing}{tb}{lst}{}
\floatname{listing}{Listing}
\title{Lost in Time? A Meta-Learning Framework for Time-Shift-Tolerant\\ Physiological Signal Transformation}

\author{
    Qian Hong\textsuperscript{\rm 1}\equalcontrib,
    Cheng Bian\textsuperscript{\rm 4}\equalcontrib,
    Xiao Zhou\textsuperscript{\rm 1,2,3}\thanks{Corresponding author.},
    Xiaoyu Li\textsuperscript{\rm 4},
    Yelei Li\textsuperscript{\rm 4},
    Zijing Zeng\textsuperscript{\rm 4}
}
\affiliations{
    \textsuperscript{\rm 1}Gaoling School of Artificial Intelligence, Renmin University of China, Beijing, China\\
    \textsuperscript{\rm 2}Beijing Key Laboratory of Research on Large Models and Intelligent Governance\\  
    \textsuperscript{\rm 3}Engineering Research Center of Next-Generation Intelligent Search and Recommendation, MOE\\
    \textsuperscript{\rm 4}OPPO Health Lab, Shenzhen, China\\  
    \{qianhong99, xiaozhou\}@ruc.edu.cn, 
    \{biancheng, lixiaoyu5, liyelei1, zijing\}@oppo.com
}

\usepackage{bibentry}

\begin{document}

\maketitle

\begin{abstract}
Translating non-invasive signals such as photoplethysmography (PPG) and ballistocardiography (BCG) into clinically meaningful signals like arterial blood pressure (ABP) is vital for continuous, low-cost healthcare monitoring. However, temporal misalignment in multimodal signal transformation impairs transformation accuracy, especially in capturing critical features like ABP peaks. Conventional synchronization methods often rely on strong similarity assumptions or manual tuning, while existing Learning with Noisy Labels (LNL) approaches are ineffective under time-shifted supervision, either discarding excessive data or failing to correct label shifts. To address this challenge, we propose \textbf{ShiftSyncNet}, a meta-learning-based bi-level optimization framework that automatically mitigates performance degradation due to time misalignment. It comprises a transformation network (\textit{TransNet}) and a time-shift correction network (\textit{SyncNet}), where \textit{SyncNet} learns time offsets between training pairs and applies Fourier phase shifts to align supervision signals. Experiments on one real-world industrial dataset and two public datasets show that \textit{ShiftSyncNet} outperforms strong baselines by 9.4\%, 6.0\%, and 12.8\%, respectively. 
The results highlight its effectiveness in correcting time shifts, improving label quality, and enhancing transformation accuracy across diverse misalignment scenarios, pointing toward a unified direction for addressing temporal inconsistencies in multimodal physiological transformation.
\end{abstract}

\begin{links}
    \link{Code}{https://github.com/HQ-LV/ShiftSyncNet}
\end{links}

\section{Introduction}

\begin{figure}[!ht]
	\centering 
     \subfloat[\label{intro-shift_reason}Multimodal signal time shift.]{\includegraphics[width=.5\columnwidth]{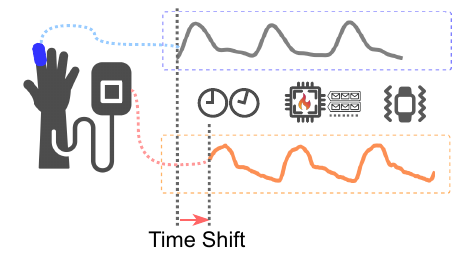}} 
      \subfloat[\label{intro-change_shift}Loss with rising time shift.]{\includegraphics[width=.5\columnwidth]{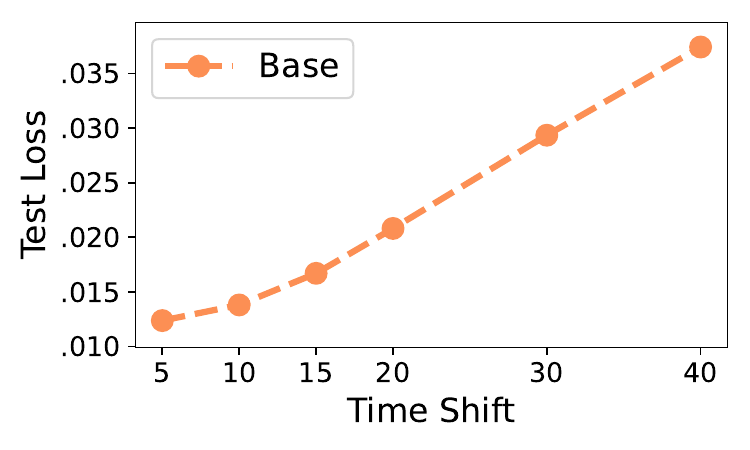}} 
    \\
    \subfloat[\label{intro-recons}Signal2Signal with time shift.]{\includegraphics[width=.5\columnwidth]{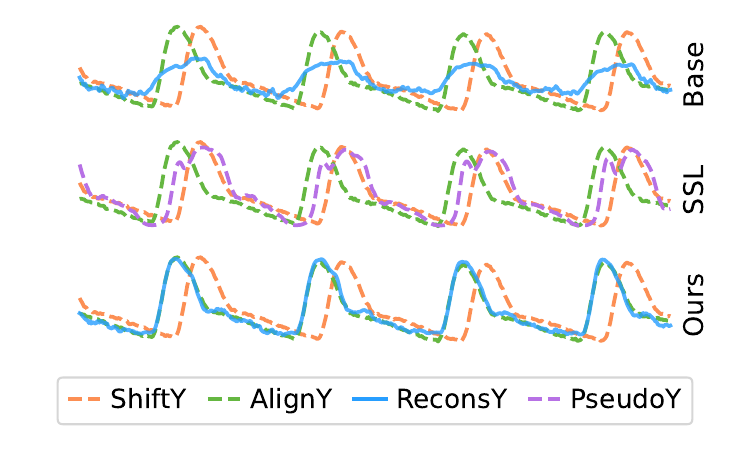}} 
    \subfloat[\label{intro-change_rate}Loss with rising corruption.]
    {\includegraphics[width=.5\columnwidth]{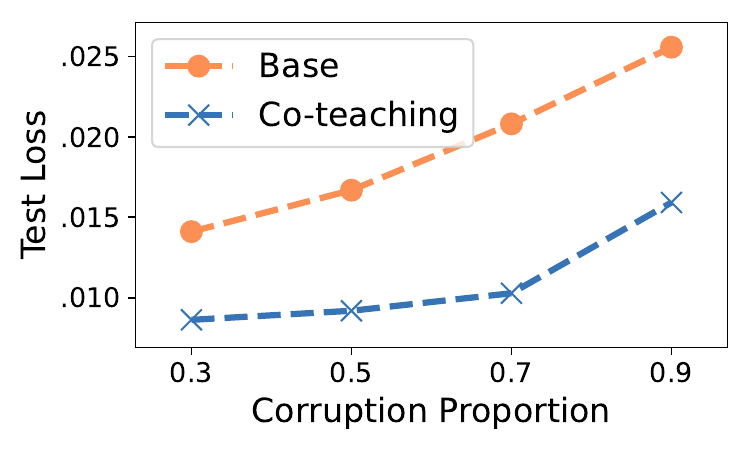}} 

\caption{Time shift between multimodal signals degrades  physiological signal transformation performance.}
 \label{intro-fig}
\end{figure}

Invasive arterial blood pressure (ABP) measurement is widely regarded as the clinical gold standard for blood pressure monitoring.
However, its measurement is generally based on invasive procedures~\cite{ogedegbe2010principles}, which carry risks of infection, discomfort, and the need for specialized medical staff, making them impractical for continuous monitoring. These limitations have motivated non-invasive alternatives such as photoplethysmography (PPG), which measures blood flow via light, and ballistocardiography (BCG), which captures cardiac mechanical activity~\cite{fortino2010ppg, misawa2022relationship,lin2025longitudinal,li2024continuous}. Converting these signals into ABP-like waveforms enables continuous, cost-effective, and wearable-friendly blood pressure monitoring.

Beyond population-level health monitoring~\cite{ijcai2018p497}, waveform transformation for personalized assessment is increasingly achieved using deep learning models, 
including CNNs~\cite{ibtehaz2022ppg2abp,cao2023incepse,chen2024improved}, GANs~\cite{golany2019pgans,sarkar2021cardiogan}, and Transformers~\cite{lan2023performer,yuan2024catransformer}. However, temporal misalignment between source and target signals remains a critical unresolved challenge.
Fig.~\ref{intro-shift_reason} illustrates potential causes of time shifts between multimodal physiological signals. For instance, PPG signals  finger-clip sensors and ABP signals  invasive brachial arterial catheters are collected by different devices and  subject to factors such as sensor-clock asynchrony, system scheduling delays, device placement variability, and firmware or driver faults, all of which can introduce temporal misalignment.
To illustrate the impact of time shifts on waveform transformation, Fig.~\ref{intro-change_shift},~\ref{intro-recons},~\ref{intro-change_rate} present PPG-to-ABP results  the \textit{Base model}, an InceptionTime~\cite{ismail2020inceptiontime} network trained on corrupted pairs with misaligned physiological signals. As the magnitude of the time shift increases (Fig.~\ref{intro-change_shift}) or the proportion of corrupted samples grows (Fig.~\ref{intro-change_rate}), the \textit{Base model}’s performance declines markedly. To visualize this effect, Fig.~\ref{intro-recons} (top row) shows the predicted ABP waveforms, where ``ShiftY", ``AlignY", and ``ReconsY" denote the misaligned labels, aligned labels, and model outputs. The results show degraded transformation performance, especially in capturing ABP peaks which is critical for hypertension diagnosis.
Although existing biomedical signal synchronization methods~\cite{xiao2022time,goodwin2023truth,boljanic2023comparison,eleveld2024haemosync} can alleviate such issues, they generally assume similar waveform shapes or require manual tuning, which limits their broader applicability. This motivates our first research question \textbf{(RQ1)}: \textit{How can we automatically mitigate the performance degradation caused by sensor time misalignment in physiological signal transformations?}

A closely related direction to signal transformation under sensor time misalignment is \textit{Learning with Noisy Labels} (LNL)~\cite{song2022learning}, where class label noise often stems  non-expert annotations or automated labeling on crowdsourcing platforms. Existing LNL methods are typically categorized into \textbf{sample selection}, which filters out suspected noisy samples~\cite{han2018co}, and \textbf{label correction}, which adjusts incorrect labels to improve supervision~\cite{shu2019meta}.  
As shown in Fig.~\ref{intro-change_rate}, the performance of Co-teaching~\cite{han2018co}, a representative sample selection method, degrades sharply as the corruption rate increases.
Early label correction methods, such as noise transition matrix estimation~\cite{goldberger2017training}, mainly target discrete classification labels and therefore struggle with regression tasks or complex noise like time shifts. To reduce information loss, some studies employ semi-supervised learning (SSL) to generate hard or soft pseudo-labels~\cite{zhang2017mixup,menon2020can,arazo2019unsupervised,li2020dividemix,zheltonozhskii2022contrast,xiao2022promix,zhang2024cross}. However, when applied to noisy data with over-parameterized models, SSL often produces unreliable pseudo-labels. As shown in Fig.~\ref{intro-recons}, SSL-generated soft labels (``PseudoY") may blend peaks  different time shifts, leading to misalignment with the true supervision signal. Although recent studies consider label corruption in time series~\cite{ma2023ctw,nagaraj2024learning}, they remain focused on classification and do not address time misalignment in sequence-to-sequence tasks.
In waveform transformation, sensor time misalignment introduces label noise similar to LNL but with a key difference: misaligned labels still retain valuable waveform characteristics. Existing LNL methods often overlook this property, highlighting the need for approaches that handle high label corruption while retaining useful signal information.  
This motivates our second research question (\textbf{RQ2}): \textit{How can we leverage time-shifted labels to infer the time offset and obtain a corrected, aligned supervision signal?}

In real-world scenarios, only a limited number of correctly labeled samples are typically available. This observation motivates the adoption of meta-learning~\cite{9428530} for label correction~\cite{zheng2021meta,wu2021learning}, where a separate meta-network is trained on a small, clean auxiliary dataset to correct corrupted supervision signals. 
Building on this approach, a meta-learning label correction scheme is expected to address these challenges by learning time offsets and fully utilizing the waveform features preserved in time-shifted labels.
 
To explore \textbf{RQ1}, we propose a meta-learning-based bi-level optimization framework, \textbf{ShiftSyncNet}, for waveform transformation tasks with tolerance to time-shifts. Specifically, our framework consists of two networks: the {waveform transformation network} (\textit{TransNet}), which translates source signals into target signals, and the {time-shift correction network} (\textit{SyncNet}), which generates potentially aligned supervision signals  corrupted labels. Both networks are trained simultaneously via  bi-level optimization: \textit{TransNet} is updated using pseudo-labels  \textit{SyncNet}, while \textit{SyncNet} is updated based on losses computed on a small, aligned meta-dataset. 
To address \textbf{RQ2}, \textit{SyncNet} handles inherent time offsets in training pairs by exploiting the Fourier shift property. This allows frequency-domain phase shifts to align misaligned supervision with the correct targets, increasing effective training data and improving model performance.

The contributions of this paper are as follows:
\begin{itemize}
    \item We tackle the time-shift challenge in waveform transformation using a meta-learning bi-level optimization framework, where \textit{SyncNet} corrects misaligned labels. 
    \item We propose the application of phase shifts in frequency domain to produce accurately aligned supervision signals, based on the time offsets learned by \textit{SyncNet}. 
    \item Experiments on real-world and public datasets show that our framework outperforms baselines in correcting time shifts, improving data usability, and enhancing accuracy.
\end{itemize}

\section{Related Work} 
\subsection{Physiological Signal Transformation}
Recent work has focused on reconstructing ABP wearable signals using deep learning. LSTMs~\cite{harfiya2021continuous}, CNNs~\cite{ibtehaz2022ppg2abp,pan2024robust}, and InceptionTime~\cite{cao2023incepse,chen2024improved} are widely used for modeling temporal dependencies. GANs~\cite{golany2019pgans,sarkar2021cardiogan} generate realistic waveforms but face stability issues. Transformer models~\cite{lan2023performer,yuan2024catransformer} handle long-range dependencies but may lose temporal information in periodic time series tasks due to permutation-invariant attention and tend to memorize patterns rather than learn underlying periodicity~\cite{zeng2023transformers,dong2024fan}.

\subsection{Multimodal Physiological Signal Alignment}
Various methods have been proposed to correct time offsets in multimodal physiological signals. Dynamic Time Warping (DTW)~\cite{hong2017solving,liu2019generalized,jiang2020eventdtw} aligns time series of different lengths, while cross-correlation methods~\cite{xiao2022time} align signals by finding maximum correlation. Shared physiological artifacts (e.g., heartbeat, motion, coughing)~\cite{goodwin2023truth}, can also support synchronization. However, these methods are effective mainly for signals with similar waveforms or inherent alignment  (e.g., smartwatch ECG vs. patch ECG)~\cite{goodwin2023truth}, and perform poorly with large waveform differences or physiological delays (e.g., PPG and ABP). Empirical approaches~\cite{eleveld2024haemosync} incrementally detect offsets but rely on manual tuning and domain knowledge, limiting scalability.

\subsection{Learning with Noisy Labels}
Many LNL methods~\cite{song2022learning} begin by identifying and eliminating~\cite{han2018co} or down-weighting~\cite{shu2019meta} noisy samples.
However, their effectiveness drops as the proportion of corrupted labels increases because fewer samples remain usable. To overcome this, label-correction methods have been developed. Noise-transition matrix estimation~\cite{goldberger2017training} struggles under complex noise types. SSL-based approaches, such as regularization techniques~\cite{zhang2017mixup,menon2020can} or dual-model frameworks~\cite{arazo2019unsupervised,li2020dividemix,zheltonozhskii2022contrast,xiao2022promix,zhang2024cross}, rely on model predictions for soft or hard relabeling, but these pseudo-labels often become unreliable due to overfitting on noisy data. Meta-learning~\cite{9428530} instead corrects supervision using a separate meta-network trained on a small clean meta-dataset, and recent work reduces its computational cost with gradient approximation techniques~\cite{zheng2021meta,wu2021learning}.

Compared to manual signal time synchronization and classical LNL methods, \textit{ShiftSyncNet} directly learns time offsets and automatically generates aligned supervision, improving data usability and enabling time-shift-tolerant signal transformation.

\section{Problem Statement}
We formulate the problem of waveform transformation affected by sensor time misalignment. Given a source signal time series $x$, the objective is to learn a waveform transformation network $ f_\theta$  parameterized by $\theta$, which maps  $x$  to the target signal series $y:y = f_\theta(x)$.
In scenarios where there is a unknown temporal shift $s$ between the source and target signals, two types of datasets are utilized: temporally misaligned training dataset:  $D'= \{(x, y')\}^N$, and temporally aligned metaset:  $D = \{(x_m, y_m)\}^M$, where  $M \ll N $.
In the training dataset  $D'$, the sample pairs $ (x, y')$  exhibit unknown temporal shifts, leading to disrupted physiological correspondences between the source and target signals.

\begin{figure*}[!ht]
\centering 
    \subfloat {\includegraphics[width=.82\textwidth]{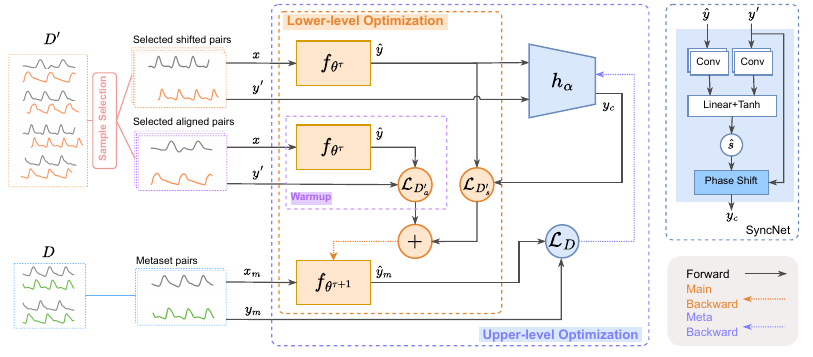}}  
\caption{Overview of \textit{ShiftSyncNet}. It follows a bi-level optimization structure: the lower level updates \textit{TransNet} $f_\theta$ to minimize the training loss on the misaligned dataset $D^{\prime}$ using labels corrected by \textit{SyncNet} $h_\alpha$, while the upper level updates \textit{SyncNet} to minimize \textit{TransNet}’s loss on the clean metaset $D$.}

 \label{framework} 
\end{figure*}

\section{Methodology}
As shown in Fig.~\ref{framework}, we propose a time-shift-tolerant meta-learning framework to address the challenge of physiological waveform transformation under time-shift interference. \textit{TransNet} $f_\theta$ is optimized as the primary objective to perform the waveform transformation task. To obtain corrected supervision signals, we introduce a meta-network, time-shift correction network (\textit{SyncNet}) $h_\alpha$, whose optimization serves as the meta-objective. \textit{SyncNet} learns the time-shift $s$  misaligned training samples $(x, y')$ in $D^{\prime}$ and applies phase shifting in frequency domain to generate aligned supervision signals $y_c$. These corrected signals provide more accurate supervision for the \textit{TransNet}.
Within this meta-learning framework, we jointly optimize both  $f_\theta$ and $h_\alpha$. 

This section begins by formulating the bi-level optimization problem, followed by detailed introductions to the meta-gradient approximation method, the \textit{SyncNet} architecture, and the \textit{sample-selection-based training strategy}.

\subsection[Problem Formulation]{bi-level Optimization Problem Formulation} 
Intuitively, if \textit{SyncNet} $h_\alpha$ provides high-quality corrected and aligned supervision for $D^{\prime}$, \textit{TransNet} $f_\theta$ should obtain low loss on the aligned metaset $D$. This leads to the following bi-level optimization problem:
\begin{equation}
\min_{\alpha}\ \mathcal{L}_{D}(\theta_\alpha^*)
\quad\text{s.t.}\quad
\theta_\alpha^*=\arg\min_{\theta}\mathcal{L}_{D'}(\alpha,\theta).
\end{equation}
We specify the upper- and lower-level objectives as:
\begin{align}
\mathcal{L}_{D}({\theta_\alpha^*}) & \triangleq \mathbb{E}_{(x_m,y_m)\in {D}} \ell(y_m, f_{\theta_\alpha^*}(x_m)), \\ 
\mathcal{L}_{D'}(\alpha,\theta) & \triangleq \mathbb{E}_{(x,y')\in {D'}}  \ell (h_\alpha(f_\theta(x),y'), f_\theta(x)),
\label{eq:lower_opt}
\end{align}
\noindent where the upper-level optimization objective $\mathcal{L}_{D}({\theta_\alpha^*})$ is to adjust the meta-network parameters $\alpha$ to minimize the loss of $f_\theta$ on the metaset $D$; the lower-level optimization objective $\mathcal{L}_{D'}(\alpha,\theta)$ is to adjust the parameters $\theta$ to minimize the training loss of $f_\theta$ on $D^{\prime}$, where the labels in $D^{\prime}$ are corrected by $h_\alpha$. $\ell$ represents loss function for waveform transformation, such as Mean Squared Error (MSE) loss.

If obtaining optimal parameters $\theta^*$ for every meta-parameter $\alpha$ were required, iterative gradient-based optimization would become computationally prohibitive. Therefore, instead of precisely solving for $\theta^*$ at every $\alpha$, we adopt a widely used alternative: approximating  $\theta^*$ after performing k-step gradient descent (GD) updates on $\theta$.

\subsection{K-step GD Lookahead Meta-Gradient} 

\noindent \textbf{One-step GD Meta-Gradient Approximation.}
We start by discussing the one-step approximation for meta-parameter, i.e., when $k = 1$. Given meta-parameter $\alpha$, the optimal $\theta^*$ can be approximated by performing one-step GD update:
\begin{align}
\theta^* = \theta^{t+1} \approx \theta^{t} - \eta \nabla_{\theta} \mathcal{L}_{D^{\prime}} (\alpha, \theta^{t})  \  , 
\end{align}
\noindent where $\theta^t$ denotes the current parameters, $\theta^{t+1}$ the updated parameters, and $\eta$ the learning rate for updating $\theta$.

The gradient of $\alpha$ is calculated via the chain rule:
\begin{align}
 \frac{\partial {\mathcal{L}_{D}}{({\theta}^{*})}}{\partial \alpha} 
=\frac{\partial {\mathcal{L}_{D}}{({\theta}^{t+1})}}{\partial {\theta}^{t+1}} \frac{\partial {\theta}^{t+1}}{\partial \alpha} 
= -  {\eta} g_{{\theta}^{t+1}}H_{{\theta},\alpha}^{t}  \  ,
\label{eq:one_step_meta_gradient}
 \end{align}
where $g_{\theta^{t+1}}$ represents the gradient of the training loss on the metaset $D$, and $\frac{\partial \theta^{t+1}}{\partial \alpha}$ is computed as follows:
\begin{align}
\frac{\partial {\theta}^{t+1}}{\partial \alpha} 
&= -{\eta}\frac{\partial}{\partial \alpha}\nabla_{\theta} {\mathcal{L}_{D'}}(\alpha,{\theta}^{t}) 
 = -{\eta} H_{{\theta},\alpha}^t   \  ,
\end{align} 
\noindent where $H_{{\theta},\alpha}^t = \nabla_{{\theta},\alpha} {  \mathcal{L}_{D'}(\alpha,{\theta}^{t})} $.

\noindent \textbf{K-step GD Meta-Gradient Approximation.}
For $k > 1$, the backbone model updates $k$ steps for each meta-network update. For $\tau$ where $t-\!k+1 \leq \tau \leq t$,  the meta-parameters remain unchanged, i.e., $\alpha^t\!=\!\alpha^{t-1}\!= \dots = \alpha^{t-k+1}$. Thus, the parameters $\theta$ for the previous $k$ steps depend on $\alpha$: 
 {\small{
\begin{equation}
\frac{\partial {\theta}^{{\tau}+1}}{\partial \alpha}  
\!=\! \frac{\partial}{\partial \alpha} ({\theta}^{{\tau}}\! -\! \eta \nabla_{\theta} {\mathcal{L}_{D'}}(\alpha,{\theta}^{{\tau}}(\alpha))) 
\! \approx \! (1\! - \! {\eta} )\frac{\partial {\theta}^{{\tau}}}{\partial \alpha} \!-\! {\eta} H_{\theta,\alpha}^{{\tau}},
\label{eq:partial_theta_alpha}
\end{equation}}}
\noindent where we approximate $H_{\theta,\theta}^\tau=\nabla_{{\theta},{\theta}}   \mathcal{L}_{D'}(\alpha,{\theta}^{{\tau}})\!\approx I$.
Expanding the recursion gives:
{\small{
\begin{equation}
\frac{\partial {\theta}^{\tau+1}}{\partial \alpha} 
= -  {\eta} H_{{\theta},\alpha}^{\tau} - {\eta}\sum_{j=1}^{k-1}(1 - {\eta} )^jH_{{\theta},\alpha}^{\tau-j}.
\label{eq:partial_theta_alpha_exp_recursive}
\end{equation}
}}
Substituting above $\frac{\partial {\theta}^{\tau+1}}{\partial \alpha} $ to solve for meta-gradient gives:
\begin{equation}
 \frac{\partial {\mathcal{L}_{D}}{({\theta}^{\tau+1})}}{\partial \alpha}  
\!=\! g_{{\theta}^{\tau+1}} \!  \frac{\partial {\theta}^{\tau+1}}{\partial \alpha} 
 \! \approx \!  -  {\eta}g_{{\theta}^{\tau+1}} H_{{\theta},\alpha}^{\tau} \!+\!  \lambda \frac{\partial {\mathcal{L}_{D}}{({\theta}^{\tau})}}{\partial \alpha}, 
\label{eq:k_step_meta_gradient}
\end{equation}
\noindent  where  $\lambda = 1 - {\eta} $. 
The first term in Eq.~\ref{eq:k_step_meta_gradient} represents the current meta-gradient, consistent with the case $k=1$ in Eq.~\ref{eq:one_step_meta_gradient}. The second term approximates the cumulative gradients  the previous $k-1$ steps with a discount factor $\lambda$, 
requiring only the latest meta-gradient to be stored.

The meta-gradient for current step can be computed as: 
 \begin{align}
    \eta g_{{\theta}^{\tau+1}}   H_{\theta,\alpha}
 = \nabla_{\alpha} (\eta \nabla_{{\theta}}^{\top} \mathcal{L}_{D'} (\alpha, {\theta}^{\tau})   \nabla_{{\theta}} \mathcal{L}_{D} ({\theta}^{\tau+1})).
\end{align}
 
For proofs of Eq.~\ref{eq:partial_theta_alpha},~\ref{eq:partial_theta_alpha_exp_recursive} and~\ref{eq:k_step_meta_gradient}, refer to App.~\ref{sec:proof}.

\subsection{Correct Shifted Labels via Time Shifting Property of Fourier Transform} 
Returning to the task in this paper, our objective is to generate a corrected supervision signal $y$ based on the output  $\hat{y} = f(x)$   \textit{TransNet} and the label $y^{\prime}$. A naive design for the \textit{SyncNet} involves applying two separate 1D-CNNs to extract features  $\hat{y}$  and $y^{\prime}$, followed by producing a corrected signal $y_c$. However, this approach has proven ineffective in learning the correct label, 
as it underemphasizes the richer supervisory information contained in $y'$.

Recalling the temporally misaligned sample pair $(x, y^{\prime})$, we hypothesize that in the later stages of training, if $f_\theta$ becomes sufficiently accurate, 
its output $\hat{y}$  and the label $y^{\prime}$ should differ only by the time-offset $s$, with their underlying waveforms remaining highly similar.
Even in the early training stages, when the waveform characteristics of $\hat{y}$ and $y^{\prime}$ still differ significantly, we can still estimate the time offset $s$ by exploiting the characteristics of the two periodic signals. Therefore, we modify the \textit{SyncNet}’s task: $h_\alpha$ is trained to learn the $s$  $ \hat{y}$  and $y^{\prime}$. By trimming a segment of length $s$  the beginning of $\hat{y}$ and the end of $y'$, or vice versa, the remaining middle parts are the aligned segments, which can be used for loss calculation. However, the rounding operation applied to $s$ and the slicing operation on $\hat{y}$ make the loss non-differentiable, making it impossible for \textit{SyncNet} to learn.
To ensure that the learned $s$ for label-correction contributes to the loss calculation and enables gradient backpropagation, we utilize the following time shifting property of Fourier transforms.

\begin{statement}
The time shifting property of Fourier transform states that shifting a signal $y'$ by $t_0$ in time domain introduces a linear phase shift in frequency spectrum with a slope of $-\omega t_0$.  
Therefore, if
\begin{align}
    y'(t) \stackrel{FT}{\leftrightarrow} Y'(\omega),
\end{align}
\noindent then based on time-shifting property of Fourier transform,
\begin{align}
y'(t - t_0) \stackrel{FT}{\leftrightarrow} e^{-j\omega t_0}Y'(\omega).
\end{align}
\end{statement}

Therefore, after \textit{SyncNet} learns the time-offset $s$, we apply a \textit{Fourier transform} to $y^{\prime}$ to obtain $Y^{\prime}\!= \text{FFT}(y^{\prime})$. In the frequency domain, a linear phase shift is applied as $
Y_c = Y^{\prime} \cdot e^{-j 2 \pi f s}$, followed by an inverse \textit{Fourier transform} to obtain the corrected time-domain signal $y_c = \text{IFFT}(Y_c)$. Here, $s$ is embedded in the exponential term of Fourier transform result, making it differentiable.

\subsection{Sample-selection-based Training Strategy}
To better exploit the metaset, we initialize the backbone pretrained on it. As shown in prior work~\cite{zhang2021understanding,han2018co}, deep networks first learn clean, simple patterns, so early-stage losses can separate aligned  corrupted samples. However, as training proceeds, shifted samples increasingly disrupt learning. Thus, we select only likely aligned samples for warmup in the first $e$ epochs, using the following loss:
\begin{align}
\mathcal{L}_{D'_a}(\theta) \triangleq \mathbb{E}_{(x,y') \in D'_a} \ell(y', f_\theta(x)),
\label{eq:selected_aligned_loss}
\end{align} 
\noindent where ${D'_a}$ refers to the $1-r$ proportion of low-loss samples selected using the Co-teaching~\cite{han2018co} threshold, with the remaining samples discarded.

In later epochs, we also partition each batch into potentially aligned and misaligned samples, computing $\mathcal{L}_{D'_a}$ with $y'$ for selected aligned ones and using corrected supervision $y_c$  $h_\alpha$ for selected misaligned ones:
\begin{align}
\mathcal{L}_{D'_s}(\alpha,\theta) \triangleq \mathbb{E}_{(x,y') \in D'_s} \ell(h_\alpha(f_\theta(x), y'), f_\theta(x)).
\label{eq:selected_shifted_loss}
\end{align}

The soft loss for updating $f_\theta$ has two terms weighted by the ratios of selected aligned and misaligned samples:
\begin{align} 
\mathcal{L}_{D'_{\text{soft}}} = \beta \mathcal{L}_{D'_a} + (1 - \beta) \mathcal{L}_{D'_s}, \ \beta = \frac{|D'_a|}{|D'_a| + |D'_s|}.
\label{eq:lower_opt_soft}
\end{align}

\section{Experiments}
 
\subsection{Datasets} 
We evaluate our approach on three datasets listed in Table~\ref{tab:statistical-description}. The first is real-world industrial data  OPPO Health Lab\footnote{https://www.oppo.com/en/}, a provider of smart healthcare solutions. To assess generalizability, we also use two public datasets: VitalDB~\cite{lee2022vitaldb} and MIMIC II~\cite{saeed2011multiparameter}.

    
 
{
\setlength{\tabcolsep}{1mm}
\begin{table}[ht!]
    \centering
    
     { \small
        \begin{tabular}{lccc}
            \toprule
            Datasets & VitalDB & MIMIC II & OML \\
            \midrule
            \#Subjects & 144 & 942 & 182 \\
            \#Segments & 517,200&364,774& 11,529 \\
            SBP(mmHg) & $120.38\pm19.63$ & $134.19\pm22.93$ &$115.09\pm12.24$  \\
            DBP(mmHg) & $66.14\pm11.45$ & $60.21\pm12.60$ &  $74.78\pm6.80$ \\
            \bottomrule
        \end{tabular}
    }
    \caption{Dataset statistical description.}  
    \label{tab:statistical-description}
\end{table}
}

{
\setlength{\tabcolsep}{0.1mm}
\begin{table*}[ht!]
    \centering 
    { \small
    \begin{tabular}{ c   c c c   c c c  c c c }
        \toprule
       \multirow{2}{*}{Methods}  & \multicolumn{3}{ c }{VitalDB} & \multicolumn{3}{ c  }{MIMIC II} & \multicolumn{3}{ c }{OML} \\
        \cmidrule(lr){2-4} \cmidrule(lr){5-7}  \cmidrule(lr){8-10} 
        & MSE $\downarrow$ & PRD $\downarrow$& MAE $\downarrow$& MSE $\downarrow$& PRD $\downarrow$& MAE $\downarrow$& MSE $\downarrow$& PRD $\downarrow$ & MAE $\downarrow$\\
\midrule   
 SwinTransformer 
& ${0.065_{\pm0.001}}$ & ${5.492_{\pm0.027}}$ & ${0.205_{\pm0.002}}$
& ${0.070_{\pm0.001}}$ & ${5.700_{\pm0.026}}$ & ${0.214_{\pm0.001}}$
& ${0.055_{\pm0.001}}$ & ${5.522_{\pm0.026}}$ & ${0.187_{\pm0.001}}$ \\
 ResNet  
& ${0.021_{\pm0.000}}$ & ${3.153_{\pm0.018}}$ & ${0.109_{\pm0.001}}$
& ${0.031_{\pm0.000}}$ & ${3.798_{\pm0.010}}$ & ${0.136_{\pm0.000}}$
& ${0.035_{\pm0.001}}$ & ${4.437_{\pm0.046}}$ & ${0.144_{\pm0.001}}$ \\
 InceptionTime  
& ${0.021_{\pm0.000}}$ & ${3.100_{\pm0.016}}$ & ${0.109_{\pm0.001}}$
& ${0.031_{\pm0.000}}$ & ${3.771_{\pm0.009}}$ & ${0.135_{\pm0.000}}$
& ${0.036_{\pm0.001}}$ & ${4.466_{\pm0.033}}$ & ${0.144_{\pm0.001}}$ \\
\midrule
 MW-Net  
& ${0.019_{\pm0.001}}$ & ${2.972_{\pm0.058}}$ & ${0.104_{\pm0.002}}$
& ${0.028_{\pm0.000}}$ & ${3.591_{\pm0.021}}$ & ${0.127_{\pm0.001}}$
& ${0.035_{\pm0.000}}$ & ${4.393_{\pm0.025}}$ & ${0.140_{\pm0.001}}$ \\
 Co-teaching  
& ${0.010_{\pm0.000}}$ & ${2.167_{\pm0.021}}$ & {${\boldsymbol{0.065}_{\pm0.001}}$}
& ${0.019_{\pm0.000}}$ & ${2.945_{\pm0.015}}$ & {${\boldsymbol{0.082}_{\pm0.000}}$}
& ${0.025_{\pm0.000}}$ & ${3.760_{\pm0.033}}$ & ${0.109_{\pm0.001}}$ \\
\midrule 
 U-correction  
& ${0.019_{\pm0.000}}$ & ${2.985_{\pm0.000}}$ & ${0.103_{\pm0.000}}$
& ${0.028_{\pm0.000}}$ & ${3.623_{\pm0.000}}$ & ${0.128_{\pm0.000}}$
& ${0.039_{\pm0.000}}$ & ${4.622_{\pm0.000}}$ & ${0.156_{\pm0.000}}$ \\
 DivideMix  
& ${0.017_{\pm0.001}}$ & ${2.779_{\pm0.097}}$ & ${0.089_{\pm0.007}}$
& ${0.022_{\pm0.001}}$ & ${3.167_{\pm0.095}}$ & ${0.100_{\pm0.004}}$
& ${0.035_{\pm0.000}}$ & ${4.379_{\pm0.012}}$ & ${0.142_{\pm0.001}}$ \\
 MSLC  
& ${0.022_{\pm0.001}}$ & ${3.170_{\pm0.085}}$ & ${0.112_{\pm0.004}}$
& ${0.031_{\pm0.000}}$ & ${3.806_{\pm0.016}}$ & ${0.137_{\pm0.001}}$
& ${0.036_{\pm0.001}}$ & ${4.450_{\pm0.075}}$ & ${0.144_{\pm0.004}}$ \\
        
 MLC  
& ${0.014_{\pm0.000}}$ & ${2.516_{\pm0.014}}$ & ${0.083_{\pm0.000}}$
& ${0.021_{\pm0.000}}$ & ${3.146_{\pm0.028}}$ & ${0.100_{\pm0.002}}$
& ${0.034_{\pm0.001}}$ & ${4.338_{\pm0.056}}$ & ${0.135_{\pm0.002}}$ \\
  C2MT  
& ${0.018_{\pm0.001}}$ & ${2.860_{\pm0.123}}$ & ${0.089_{\pm0.004}}$
& ${0.020_{\pm0.000}}$ & ${3.071_{\pm0.014}}$ & ${0.094_{\pm0.002}}$
& ${0.034_{\pm0.001}}$ & ${4.341_{\pm0.054}}$ & ${0.135_{\pm0.002}}$ \\
\midrule
 Ours  
& {${\boldsymbol{0.009}_{\pm0.000}}$} & {${\boldsymbol{2.097}_{\pm0.015}}$} & ${0.066_{\pm0.001}}$
& {${\boldsymbol{0.016}_{\pm0.000}}$} & {${\boldsymbol{2.755}_{\pm0.009}}$} & ${0.083_{\pm0.000}}$
& {${\boldsymbol{0.023}_{\pm0.001}}$} & {${\boldsymbol{3.572}_{\pm0.038}}$} & {${\boldsymbol{0.108}_{\pm0.001}}$} \\
        \bottomrule
    \end{tabular}
    }
    \caption{Performance comparison of our method with baselines. \normalfont{\textbf{Bold} indicates the best results. $\downarrow$ means lower is better.}}
    \label{tab:overall-performance}
\end{table*}
}

\noindent\textbf{OML.}   
Our industrial dataset OML~\cite{bian2024constraint}, collected  180+ subjects via specialized wearables, includes 125 Hz BCG and PPG signals. We apply Butterworth bandpass filtering (BCG) and derivative-based mean filtering (PPG), segment signals into 6.14~s slices (768 points) and split subject into train/valid/test sets in an 8:1:1 ratio.

\noindent\textbf{VitalDB.}  
VitalDB~\cite{lee2022vitaldb} contains 6,388 PPG, ECG, and ABP records  ICU patients at Seoul National University Hospital. We adopt the cleaned version in~\cite{wang2023pulsedb}, downsampled to 125~Hz and split by subject into training and test sets. Signals are segmented into 6.14~s slices, with 10\% of training data for validation.

\noindent \textbf{MIMIC II.}  
A subset of MIMIC-II~\cite{saeed2011multiparameter}  PhysioNet~\cite{goldberger2000physiobank}, collected at Beth Israel Deaconess Medical Center, using the preprocessed version by Kachuee et al.~\cite{kachuee2015cuff,kachuee2016cuffless}, includes 12,000 PPG, ECG, and ABP records  942 ICU patients at 125 Hz. Signals are sliced into 6.14~s windows and split into train/valid/test sets (8:1:1).

\subsection{Experimental Settings}
\noindent \textbf{Time-shift Settings.} 
As real time-shifts are often unavailable, we simulate it by injecting artificial shifts into the training data, enabling controlled evaluation without ground-truth alignment and improving robustness by exposing the model to more diverse, complex misalignments.
To simulate complex time-shift scenarios, we vary: 
(1) shift magnitude: each corrupted sample is randomly shifted to left or right by $s$ points ($s\!\leq\! S$), with $S \in \{5, 10, 15, 20, 30, 40\}$;  
(2) corruption proportion: a fraction $r$ of samples is shifted, with $r \in \{0.3, 0.5, 0.7, 0.9\}$.  
These settings enable evaluation of method’s robustness under varying time-shift conditions.
 
\noindent \textbf{Evaluation Metrics.}
We use Mean Squared Error (MSE), Percent Root Difference (PRD), and Mean Absolute Error (MAE) for waveform transformation evaluation.

\begin{figure*}[!ht]
\centering 
    \subfloat[\label{recons_compare_s0}$s=0$]{\includegraphics[width=.32\textwidth]{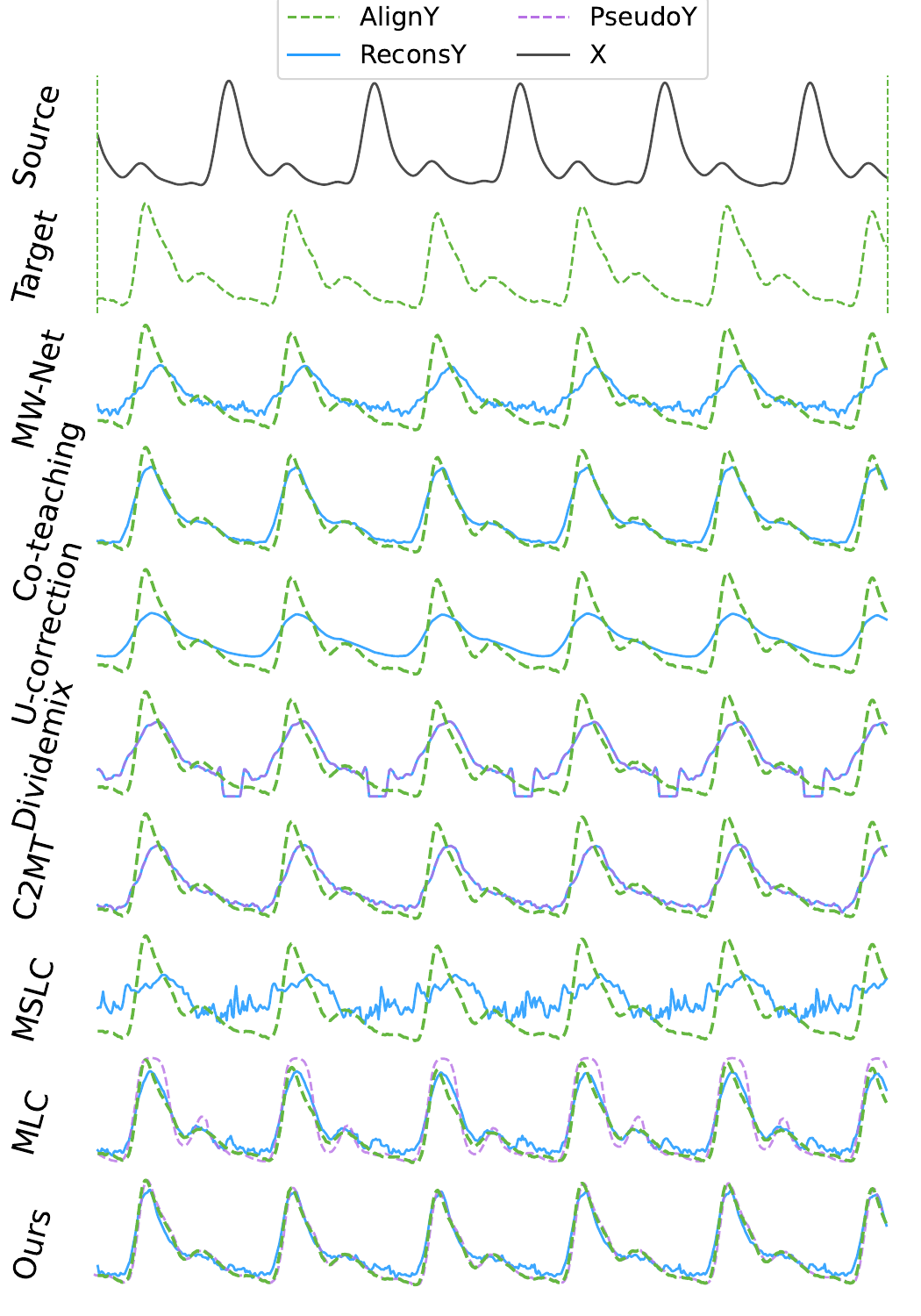}} 
    \subfloat[\label{recons_compare_s-10}$s=-10$]{\includegraphics[width=.32\textwidth]{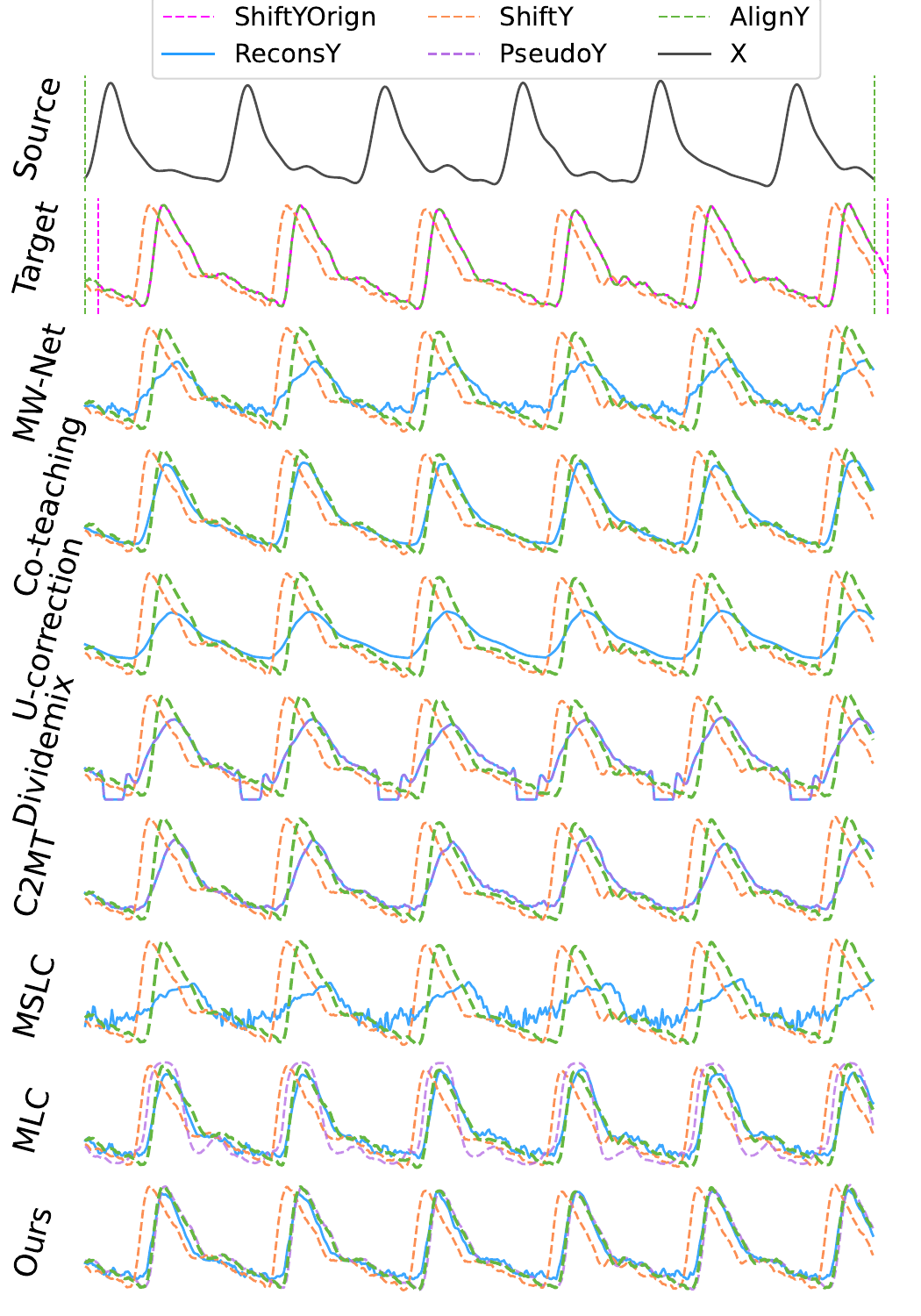}} 
    \subfloat[\label{recons_compare_s20}$s=20$]{\includegraphics[width=.32\textwidth]{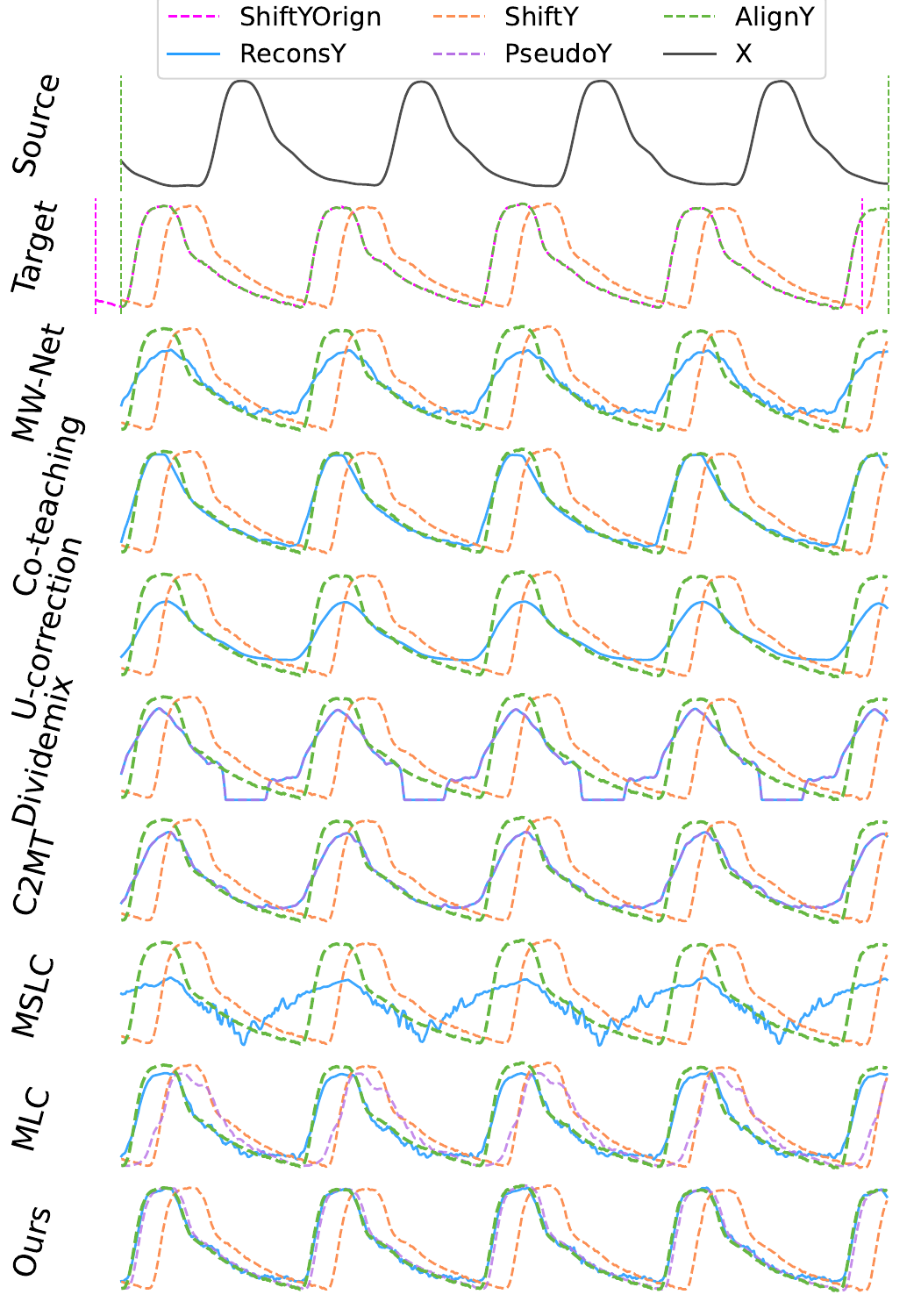}}
\caption{Signal transformation visualization under varying time-shift conditions. \normalfont{``Shift" indicates supervision with time-shift ${s}$; ``ReconsY" and ``PseudoY" represent predictive signals and pseudo-labels, respectively. For clarity, we also show the true original position of ``ShiftY" as ``ShiftYOrigin" and the input-aligned target as ``AlignY", both unknown during training.}}
 \label{recons_compare}
\end{figure*}

\noindent \textbf{Baselines.} 
We use three baseline categories: basic backbone models (Swin-Transformer~\cite{liu2021swin}, ResNet~\cite{he2016deep}, InceptionTime~\cite{ismail2020inceptiontime}), sample selection methods (MW-Net~\cite{shu2019meta}, Co-teaching~\cite{han2018co}), and label correction methods (U-correction~\cite{arazo2019unsupervised}, DivideMix~\cite{li2020dividemix}, MSLC~\cite{wu2021learning}, MLC~\cite{zheng2021meta} and C2MT~\cite{zhang2024cross}).

\subsection{Performance Evaluation} 
\noindent \textbf{Overall Evaluation.}
Table~\ref{tab:overall-performance} shows overall performance with $S\!=\!20$ and $r\!=\!0.7$, reporting average and standard deviation over 5 trials. 
More results on time-shift tolerance under various settings are in the App.~\ref{app-Experimental}. 
Among backbone models, InceptionTime and ResNet outperform SwinTransformer, with InceptionTime chosen for subsequent backbone due to its higher computational efficiency. Overall, our model achieves the best results on 7 of 9 metrics across three datasets and ranks second on the other two metrics. 

Compared to label correction methods, \textit{ShiftSyncNet} outperforms U-correction, DivideMix, MSLC, MLC, and C2MT, reducing MSE by 30.4\%, 19.6\%, and 32.2\% over the second-best method across the three datasets.   U-correction, DivideMix, and C2MT rely on semi-supervised strategies that depend heavily on over-parameterized backbone predictions, which are unreliable under noise. As shown in Fig.~\ref{recons_compare}, these methods generate overconfident pseudo-labels that deviate  true waveforms. MSLC combines historical predictions blended with shifted labels, which may distort soft signal waveform.   MLC, in contrast, employs an independent meta-network  to directly output corrected labels and performs better than most, but as seen in Fig.~\ref{recons_compare_s-10}, its pseudo-labels only proximate the waveform rather than provide entirely accurate supervision labels. This limitation becomes more pronounced under larger shifts (Fig.~\ref{recons_compare_s20}).

Compared to sample selection methods, \textit{ShiftSyncNet} outperforms Co-teaching and MW-Net, reducing MSE by 6.0\%, 12.8\%, and 9.4\% over second-best Co-teaching across the three datasets. MW-Net reweights all samples adaptively, which may mislead training and degrade performance. Co-teaching selects low-loss samples based on prior assumptions and performs better than MW-Net. However, 
as discussed in App.~\ref{app-Experimental}, 
it suffers from selection bias and limited usable samples, hindering its ability to capture fine-grained waveform features such as signal peaks (Fig.~\ref{recons_compare_s0},~\ref{recons_compare_s-10}).

Unlike these baselines, our method redefines the meta-network objective to predict time shift $s$ and reverse label by phase shift, producing truly-aligned and waveform-consistent supervision that enhances data usability.

\begin{figure}[ht!]
\centering 
    \subfloat[\label{visual-trainloss_track}Training loss history.]{\includegraphics[width=.45\columnwidth]{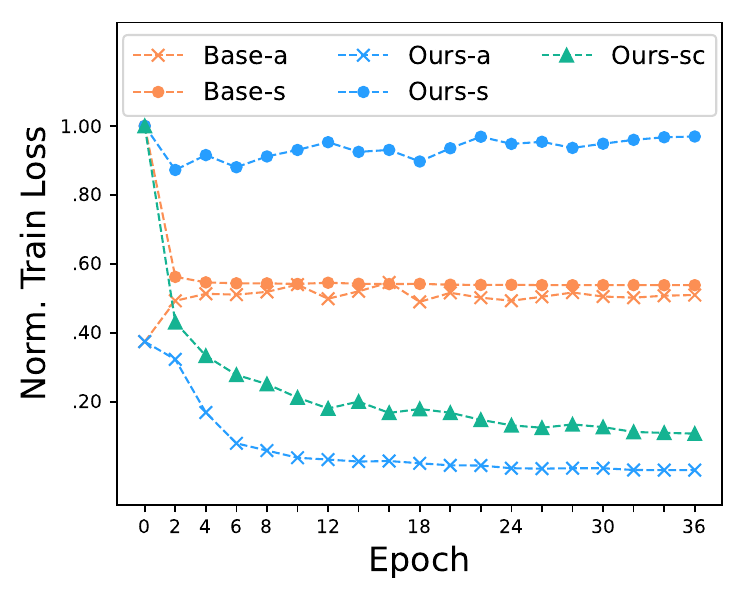}} 
    \subfloat[\label{visual-pred_shift}Time-shift prediction.]{\includegraphics[width=.45\columnwidth]{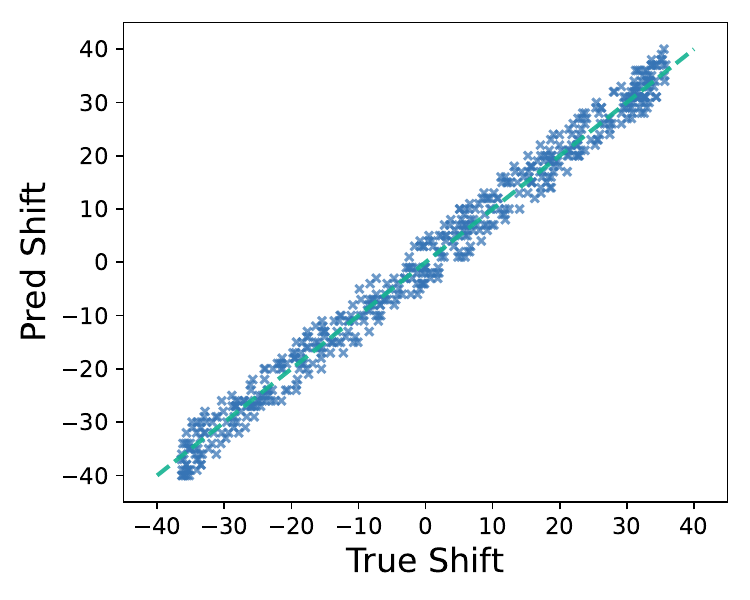}}   
\caption{Effectiveness of \textit{ShiftSyncNet}. }
\end{figure}

\begin{figure}[ht!]
\centering  
    \subfloat[\label{visual-trainloss_chosen_coteach_5}{Co-teaching, ep 5}]{\includegraphics[width=.3333\columnwidth]{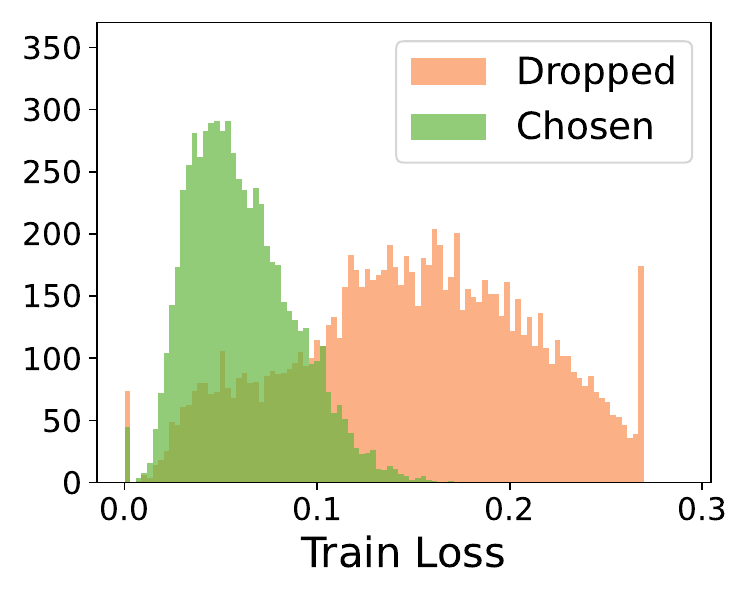}} 
    \subfloat[\label{visual-trainloss_chosen_coteach_30}Co-teaching, ep 30]{\includegraphics[width=.3333\columnwidth]{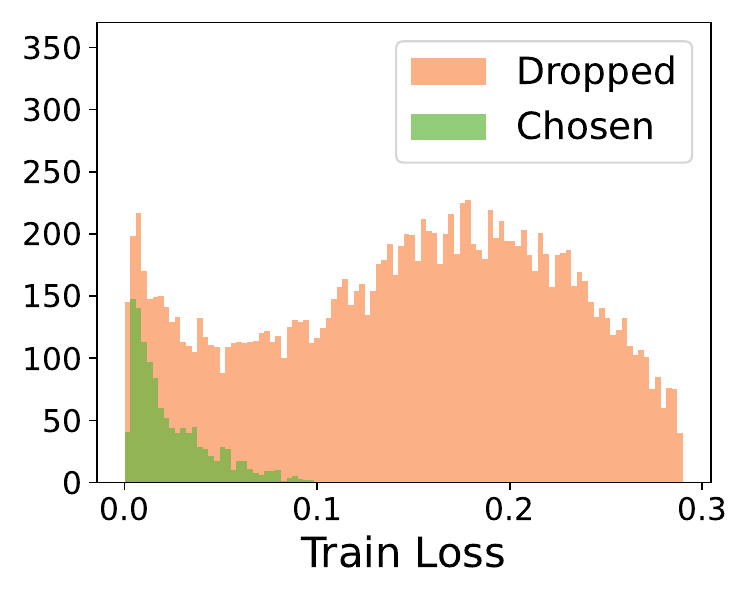}} 
    \subfloat[\label{visual-trainloss_chosen_coteach_60}Co-teaching, ep 60]{\includegraphics[width=.3333\columnwidth]{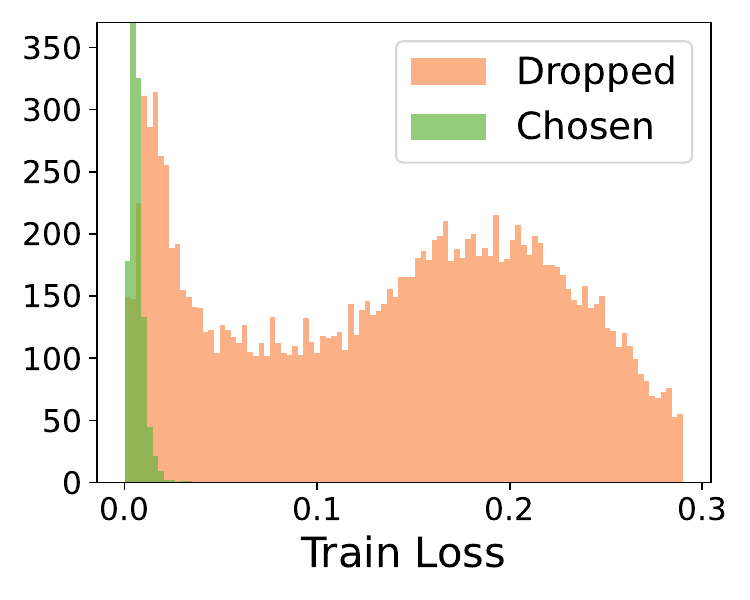}} 
    \\
    \subfloat[\label{visual-trainloss_corrected_ours_5}Ours, ep 5]{\includegraphics[width=.3333\columnwidth]{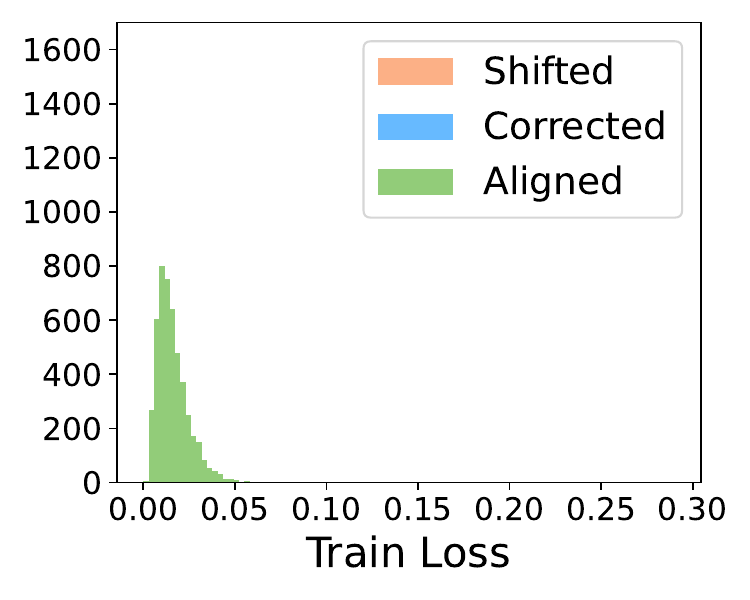}} 
    \subfloat[\label{visual-trainloss_corrected_ours_30}Ours, ep 30]{\includegraphics[width=.3333\columnwidth]{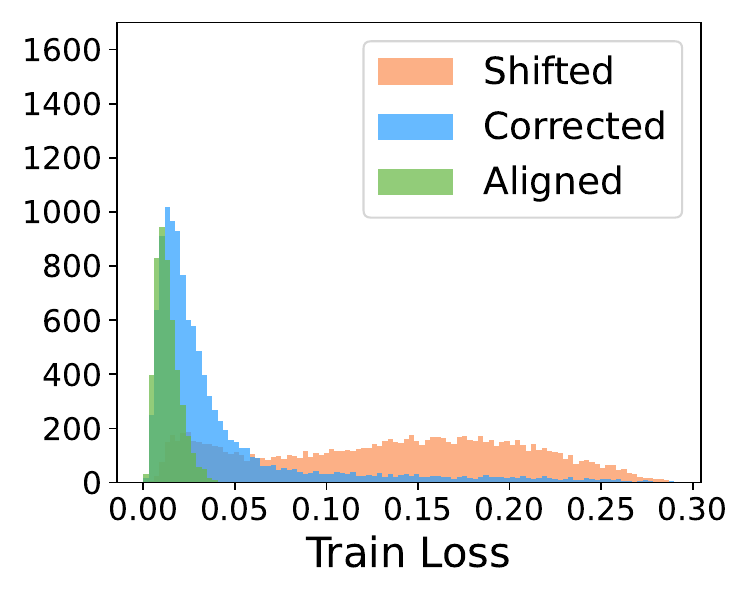}} 
    \subfloat[\label{visual-trainloss_corrected_ours_60}Ours, ep 60]{\includegraphics[width=.3333\columnwidth]{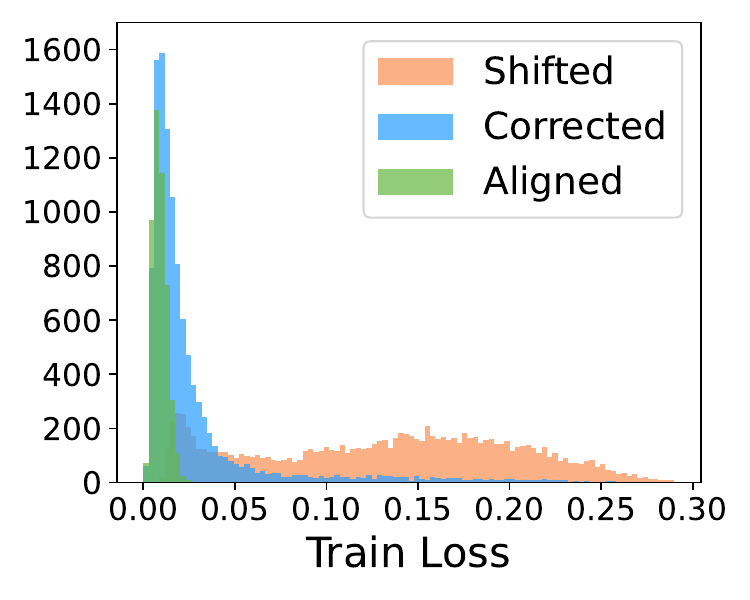}}  
\caption{Training loss distribution of aligned and shifted samples for Co-teaching and ours at epochs 5, 30, and 60.}
 \label{visual-trainloss_distribution}
\end{figure}

\noindent \textbf{Effectiveness of ShiftSyncNet.} 
We evaluate the effectiveness of \textit{ShiftSyncNet} by analyzing its handling of aligned and misaligned samples. Fig.~\ref{visual-trainloss_track} compares training loss trajectories of \textit{Base model} and \textit{ShiftSyncNet}, both using a pretrained backbone initialized on a small aligned metaset for fair comparison. Without correction, Base model notably suffers  time shifts, with even aligned sample loss (Base-a) increasing over time. In contrast, \textit{ShiftSyncNet} consistently reduces losses for aligned (Ours-a) and corrupted samples after correcting shifted labels (Ours-s $\rightarrow$ Ours-sc), indicating effective label correction and improved supervision.

To verify \textit{SyncNet}’s time-shift estimates, Fig.~\ref{visual-pred_shift} plots true time-shifts $s$ against predicted shifts $\hat{s}$, showing strong diagonal alignment $\hat{s} = s$. This enables phase-aligned pseudo-labels and improves waveform transformation.

To explore the performance gap, we compare how Co-teaching and our method handle aligned and shifted samples. 
By selecting low-loss samples and discarding high-loss ones based on a growing forgetting rate, Co-teaching gradually reduces useful data over time, hindering optimization.
In contrast, our method recovers these discarded samples by correcting the supervision of shifted pairs. As shown in Fig.~\ref{visual-trainloss_distribution}, both methods identify aligned samples (green), but Co-teaching gradually drops high-loss ones (orange in Fig.~\ref{visual-trainloss_chosen_coteach_5}, ~\ref{visual-trainloss_chosen_coteach_30}, \ref{visual-trainloss_chosen_coteach_60}). Our method corrects their labels (orange $\rightarrow$ blue in Fig.~\ref{visual-trainloss_corrected_ours_5}, ~\ref{visual-trainloss_corrected_ours_30}, ~\ref{visual-trainloss_corrected_ours_60}), significantly reducing loss.

\subsection{Downstream Task Evaluation}
To demonstrate clinical applicability, we conduct downstream systolic and diastolic blood pressure (SBP/DBP) prediction under time-shift interference on VitalDB and MIMIC-II, as shown in Table~\ref{tab:downstream}. \textit{ShiftSyncNet} achieves the best performance by mitigating time-shift effects. Compared to InceptionTime trained without time misalignment correction, it reduces SBP/DBP MAE by 80\%/72\% on VitalDB and 71\%/67\% on MIMIC-II, respectively.  
It also meets the Association for the Advancement of Medical Instrumentation (AAMI) standard (MAE $< 5$ mmHg), confirming its utility for clinical prediction and continuous monitoring.

\begin{table}[ht!]
    \centering  
\small{
    \begin{tabular}{lrrrr}
        \toprule 
        \multirow{2}{*}{Methods} & \multicolumn{2}{c}{ {VitalDB}} & \multicolumn{2}{c}{ {MIMIC II}} \\
         &  {SBP} &  {DBP} &  {SBP} &  {DBP} \\  \midrule 
        InceptionTime & 12.41 & 5.50 & 16.82 & 7.13 \\ \midrule
        MW-Net & 11.82 & 5.51 & 15.97 & 6.91 \\ 
        Co-teaching & 3.22 & 2.44 & 5.82 & 2.99 \\   \midrule
        U-correction & 12.05 &4.92 &16.08&7.01\\
        DivideMix & 6.49 & 3.37 & 7.45 & 4.93 \\
        MSLC & 13.83 & 1.93 & 19.50 & 7.48 \\  
        MLC & 5.70 & 3.60 & 7.49 & 4.91 \\  
        C2MT & 4.00 & 2.83 & 6.95 & 2.82 \\ \midrule
        Ours & \textbf{2.43} & \textbf{1.49} & \textbf{4.83} & \textbf{2.36} \\  
        Impr.& 80\% & 72\% & 71\% & 67\% \\  
         
        \bottomrule 
    \end{tabular}
    } 
    \caption{Downstream SBP/DBP prediction MAE (mmHg).  \normalfont {"Impr." denotes improvement of ours over InceptionTime.}} 
    \label{tab:downstream}
\end{table}

\subsection{Ablation Studies}
 
To evaluate \textit{sample-selection-based training strategy}, we compare our method with two variants in Table~\ref{tab:ablation}. Variant I (w/o SL) uses $\mathcal{L}_{D^{\prime}}$ in Eq.~\ref{eq:lower_opt} instead of $\mathcal{L}_{D^{\prime}_{soft}}$ in Eq.~\ref{eq:lower_opt_soft}, relying solely on pseudo-labels and ignoring aligned supervision. Variant II (w/o WU) removes warmup phase. Our method consistently outperforms both variants across time-shift settings.Variant I misses the benefits of supervision  aligned samples, while our soft loss balances aligned and corrected shifted signals for improved robustness. Variant II performs worse due to the absence of warmup, which stabilizes training by selecting low-loss samples. Combining soft loss and warmup yields superior performance.

\begin{table}[hb!]
    \centering
  {\small
    \begin{tabular}{llccc}
        \toprule 
        Dataset&Variation & $r = 0.5$ & $r = 0.7$ & $r = 0.9$ \\
        \midrule
         \multirow{3}{*}{{VitalDB}} & w/o SL     & 0.0094 & 0.0096 & 0.0098 \\
        &w/o WU     & 0.0096 & 0.0096 & 0.0096 \\
        &Ours       & \textbf{0.0093} & \textbf{0.0095} & \textbf{0.0095} \\
        \midrule 
        \multirow{3}{*}{{OML}} &w/o SL     & 0.0237 & 0.0227 & 0.0238 \\
        &w/o WU     & 0.0242 & 0.0235 & 0.0245 \\
        &Ours       & \textbf{0.0231} & \textbf{0.0226} & \textbf{0.0228} \\
        \bottomrule
    \end{tabular}
 }
 \caption{Test MSE comparison of ablation models.}
    \label{tab:ablation}
\end{table}

\section{Conclusion}
We tackle waveform transformation under time-shift interference in physiological signals with \textit{ShiftSyncNet}, a meta-learning framework comprising \textit{TransNet} for transformation and \textit{SyncNet} for time-shift correction via Fourier phase shifts. A sample-selection strategy further  enhances robustness against misaligned labels. Experiments on real-world and public datasets show that \textit{ShiftSyncNet} achieves superior accuracy and robustness under time shifts, supporting its application in continuous health monitoring.

\section{Acknowledgements}
This work was supported by the Public Computing Cloud at Renmin University of China and the Fund for Building World-Class Universities (Disciplines) at Renmin University of China.

{\small 
\bibliography{aaai2026}

\begin{thebibliography}{51}
\providecommand{\natexlab}[1]{#1}

\bibitem[{Arazo et~al.(2019)Arazo, Ortego, Albert, O’Connor, and McGuinness}]{arazo2019unsupervised}
Arazo, E.; Ortego, D.; Albert, P.; O’Connor, N.; and McGuinness, K. 2019.
\newblock Unsupervised label noise modeling and loss correction.
\newblock In \emph{International conference on machine learning}, 312--321. PMLR.

\bibitem[{Bian et~al.(2024)Bian, Li, Bi, Zhu, Lyu, Zhang, Li, and Zeng}]{bian2024constraint}
Bian, C.; Li, X.; Bi, Q.; Zhu, G.; Lyu, J.; Zhang, W.; Li, Y.; and Zeng, Z. 2024.
\newblock Constraint latent space matters: an anti-anomalous waveform transformation solution from photoplethysmography to arterial blood pressure.
\newblock In \emph{Proceedings of the AAAI Conference on Artificial Intelligence}, volume~38, 11087--11095.

\bibitem[{Boljani{\'c} et~al.(2023)Boljani{\'c}, Male{\v{s}}evi{\'c}, Vujnovi{\'c}, and Jankovi{\'c}}]{boljanic2023comparison}
Boljani{\'c}, T.; Male{\v{s}}evi{\'c}, J.; Vujnovi{\'c}, S.; and Jankovi{\'c}, M.~M. 2023.
\newblock Comparison of Time Domain Methods for Alignment of RR Signals Acquired by Different Sensor Systems.
\newblock In \emph{2023 10th International Conference on Electrical, Electronic and Computing Engineering (IcETRAN)}, 1--4. IEEE.

\bibitem[{Cao et~al.(2023)Cao, Tran, Nguyen, Nguyen, and Pham}]{cao2023incepse}
Cao, T.; Tran, N.; Nguyen, L.; Nguyen, H.; and Pham, H. 2023.
\newblock IncepSE: Leveraging InceptionTime's performance with Squeeze and Excitaion mechanism in ECG analysis.
\newblock In \emph{Proceedings of the 12th International Symposium on Information and Communication Technology}, 578--584.

\bibitem[{Chen et~al.(2024)Chen, Ke, Sun, and Wang}]{chen2024improved}
Chen, Y.; Ke, M.; Sun, Y.; and Wang, L. 2024.
\newblock An Improved InceptionTime Model for Mental Workload Assessment Based on EEG Signal.
\newblock In \emph{2024 5th International Seminar on Artificial Intelligence, Networking and Information Technology (AINIT)}, 1838--1842. IEEE.

\bibitem[{Dong et~al.(2024)Dong, Li, Tao, Jiang, Zhang, Li, Deng, Su, Zhang, and Xu}]{dong2024fan}
Dong, Y.; Li, G.; Tao, Y.; Jiang, X.; Zhang, K.; Li, J.; Deng, J.; Su, J.; Zhang, J.; and Xu, J. 2024.
\newblock Fan: Fourier analysis networks.
\newblock \emph{arXiv preprint arXiv:2410.02675}.

\bibitem[{Eleveld et~al.(2024)Eleveld, Harmsen, Elting, and Maurits}]{eleveld2024haemosync}
Eleveld, N.; Harmsen, M.; Elting, J. W.~J.; and Maurits, N.~M. 2024.
\newblock Haemosync: A synchronisation algorithm for multimodal haemodynamic signals.
\newblock \emph{Computer Methods and Programs in Biomedicine}, 108298.

\bibitem[{Fortino and Giamp{\`a}(2010)}]{fortino2010ppg}
Fortino, G.; and Giamp{\`a}, V. 2010.
\newblock PPG-based methods for non invasive and continuous blood pressure measurement: An overview and development issues in body sensor networks.
\newblock In \emph{2010 IEEE International Workshop on Medical Measurements and Applications}, 10--13. IEEE.

\bibitem[{Golany and Radinsky(2019)}]{golany2019pgans}
Golany, T.; and Radinsky, K. 2019.
\newblock Pgans: Personalized generative adversarial networks for ecg synthesis to improve patient-specific deep ecg classification.
\newblock In \emph{Proceedings of the AAAI Conference on Artificial Intelligence}, volume~33, 557--564.

\bibitem[{Goldberger et~al.(2000)Goldberger, Amaral, Glass, Hausdorff, Ivanov, Mark, Mietus, Moody, Peng, and Stanley}]{goldberger2000physiobank}
Goldberger, A.~L.; Amaral, L.~A.; Glass, L.; Hausdorff, J.~M.; Ivanov, P.~C.; Mark, R.~G.; Mietus, J.~E.; Moody, G.~B.; Peng, C.-K.; and Stanley, H.~E. 2000.
\newblock PhysioBank, PhysioToolkit, and PhysioNet: components of a new research resource for complex physiologic signals.
\newblock \emph{circulation}, 101(23): e215--e220.

\bibitem[{Goldberger and Ben-Reuven(2017)}]{goldberger2017training}
Goldberger, J.; and Ben-Reuven, E. 2017.
\newblock Training deep neural-networks using a noise adaptation layer.
\newblock In \emph{International conference on learning representations}.

\bibitem[{Goodwin et~al.(2023)Goodwin, Dixon, Mazwi, Hahn, Meir, Goodfellow, Kazazian, Greer, McEwan, Laussen et~al.}]{goodwin2023truth}
Goodwin, A.~J.; Dixon, W.; Mazwi, M.; Hahn, C.~D.; Meir, T.; Goodfellow, S.~D.; Kazazian, V.; Greer, R.~W.; McEwan, A.; Laussen, P.~C.; et~al. 2023.
\newblock The truth Hertz—synchronization of electroencephalogram signals with physiological waveforms recorded in an intensive care unit.
\newblock \emph{Physiological Measurement}, 44(8): 085002.

\bibitem[{Han et~al.(2018)Han, Yao, Yu, Niu, Xu, Hu, Tsang, and Sugiyama}]{han2018co}
Han, B.; Yao, Q.; Yu, X.; Niu, G.; Xu, M.; Hu, W.; Tsang, I.; and Sugiyama, M. 2018.
\newblock Co-teaching: Robust training of deep neural networks with extremely noisy labels.
\newblock \emph{Advances in neural information processing systems}, 31.

\bibitem[{Harfiya, Chang, and Li(2021)}]{harfiya2021continuous}
Harfiya, L.~N.; Chang, C.-C.; and Li, Y.-H. 2021.
\newblock Continuous blood pressure estimation using exclusively photopletysmography by LSTM-based signal-to-signal translation.
\newblock \emph{Sensors}, 21(9): 2952.

\bibitem[{He et~al.(2016)He, Zhang, Ren, and Sun}]{he2016deep}
He, K.; Zhang, X.; Ren, S.; and Sun, J. 2016.
\newblock Deep residual learning for image recognition.
\newblock In \emph{Proceedings of the IEEE conference on computer vision and pattern recognition}, 770--778.

\bibitem[{Hong, Park, and Baek(2017)}]{hong2017solving}
Hong, J.~Y.; Park, S.~H.; and Baek, J.-G. 2017.
\newblock Solving the singularity problem of semiconductor process signal using improved dynamic time warping.
\newblock In \emph{2017 IEEE 11th International Conference on Semantic Computing (ICSC)}, 266--267. IEEE.

\bibitem[{Hospedales et~al.(2022)Hospedales, Antoniou, Micaelli, and Storkey}]{9428530}
Hospedales, T.; Antoniou, A.; Micaelli, P.; and Storkey, A. 2022.
\newblock Meta-Learning in Neural Networks: A Survey.
\newblock \emph{IEEE Transactions on Pattern Analysis and Machine Intelligence}, 44(9): 5149--5169.

\bibitem[{Ibtehaz et~al.(2022)Ibtehaz, Mahmud, Chowdhury, Khandakar, Salman~Khan, Ayari, Tahir, and Rahman}]{ibtehaz2022ppg2abp}
Ibtehaz, N.; Mahmud, S.; Chowdhury, M.~E.; Khandakar, A.; Salman~Khan, M.; Ayari, M.~A.; Tahir, A.~M.; and Rahman, M.~S. 2022.
\newblock PPG2ABP: Translating photoplethysmogram (PPG) signals to arterial blood pressure (ABP) waveforms.
\newblock \emph{Bioengineering}, 9(11): 692.

\bibitem[{Ismail~Fawaz et~al.(2020)Ismail~Fawaz, Lucas, Forestier, Pelletier, Schmidt, Weber, Webb, Idoumghar, Muller, and Petitjean}]{ismail2020inceptiontime}
Ismail~Fawaz, H.; Lucas, B.; Forestier, G.; Pelletier, C.; Schmidt, D.~F.; Weber, J.; Webb, G.~I.; Idoumghar, L.; Muller, P.-A.; and Petitjean, F. 2020.
\newblock Inceptiontime: Finding alexnet for time series classification.
\newblock \emph{Data Mining and Knowledge Discovery}, 34(6): 1936--1962.

\bibitem[{Jiang et~al.(2020)Jiang, Qi, Wang, Bent, Avram, Olgin, and Dunn}]{jiang2020eventdtw}
Jiang, Y.; Qi, Y.; Wang, W.~K.; Bent, B.; Avram, R.; Olgin, J.; and Dunn, J. 2020.
\newblock EventDTW: An improved dynamic time warping algorithm for aligning biomedical signals of nonuniform sampling frequencies.
\newblock \emph{Sensors}, 20(9): 2700.

\bibitem[{Kachuee et~al.(2015)Kachuee, Kiani, Mohammadzade, and Shabany}]{kachuee2015cuff}
Kachuee, M.; Kiani, M.~M.; Mohammadzade, H.; and Shabany, M. 2015.
\newblock Cuff-less high-accuracy calibration-free blood pressure estimation using pulse transit time.
\newblock In \emph{2015 IEEE international symposium on circuits and systems (ISCAS)}, 1006--1009. IEEE.

\bibitem[{Kachuee et~al.(2016)Kachuee, Kiani, Mohammadzade, and Shabany}]{kachuee2016cuffless}
Kachuee, M.; Kiani, M.~M.; Mohammadzade, H.; and Shabany, M. 2016.
\newblock Cuffless blood pressure estimation algorithms for continuous health-care monitoring.
\newblock \emph{IEEE Transactions on Biomedical Engineering}, 64(4): 859--869.

\bibitem[{Lan(2023)}]{lan2023performer}
Lan, E. 2023.
\newblock Performer: A novel ppg-to-ecg reconstruction transformer for a digital biomarker of cardiovascular disease detection.
\newblock In \emph{Proceedings of the IEEE/CVF Winter Conference on Applications of Computer Vision}, 1991--1999.

\bibitem[{Lee et~al.(2022)Lee, Park, Yoon, Yang, Park, and Jung}]{lee2022vitaldb}
Lee, H.-C.; Park, Y.; Yoon, S.~B.; Yang, S.~M.; Park, D.; and Jung, C.-W. 2022.
\newblock VitalDB, a high-fidelity multi-parameter vital signs database in surgical patients.
\newblock \emph{Scientific Data}, 9(1): 279.

\bibitem[{Li, Socher, and Hoi(2020)}]{li2020dividemix}
Li, J.; Socher, R.; and Hoi, S.~C. 2020.
\newblock Dividemix: Learning with noisy labels as semi-supervised learning.
\newblock \emph{arXiv preprint arXiv:2002.07394}.

\bibitem[{Li et~al.(2024)Li, Hussein, Zhu, Sui, Li, Yang, Zeng, and Li}]{li2024continuous}
Li, X.; Hussein, R.; Zhu, G.; Sui, X.; Li, H.; Yang, X.; Zeng, Z.; and Li, Y. 2024.
\newblock Continuous Blood Pressure Monitoring and Hypertension Risk Screening Using Smart Watch.
\newblock In \emph{2024 46th Annual International Conference of the IEEE Engineering in Medicine and Biology Society (EMBC)}, 1--6. IEEE.

\bibitem[{Lin et~al.(2025)Lin, Li, Hussein, Sui, Li, Zhu, Katsaggelos, Zeng, and Li}]{lin2025longitudinal}
Lin, H.; Li, J.; Hussein, R.; Sui, X.; Li, X.; Zhu, G.; Katsaggelos, A.~K.; Zeng, Z.; and Li, Y. 2025.
\newblock Longitudinal Wrist PPG Analysis for Reliable Hypertension Risk Screening Using Deep Learning.
\newblock In \emph{ICASSP 2025-2025 IEEE International Conference on Acoustics, Speech and Signal Processing (ICASSP)}, 1--5. IEEE.

\bibitem[{Liu et~al.(2019)Liu, Yao, Wang, Plested, and Gedeon}]{liu2019generalized}
Liu, Y.; Yao, Y.; Wang, Z.; Plested, J.; and Gedeon, T. 2019.
\newblock Generalized alignment for multimodal physiological signal learning.
\newblock In \emph{2019 International Joint Conference on Neural Networks (IJCNN)}, 1--10. IEEE.

\bibitem[{Liu et~al.(2021)Liu, Lin, Cao, Hu, Wei, Zhang, Lin, and Guo}]{liu2021swin}
Liu, Z.; Lin, Y.; Cao, Y.; Hu, H.; Wei, Y.; Zhang, Z.; Lin, S.; and Guo, B. 2021.
\newblock Swin transformer: Hierarchical vision transformer using shifted windows.
\newblock In \emph{Proceedings of the IEEE/CVF international conference on computer vision}, 10012--10022.

\bibitem[{Ma et~al.(2023)Ma, Liu, Zheng, Wang, and Ma}]{ma2023ctw}
Ma, P.; Liu, Z.; Zheng, J.; Wang, L.; and Ma, Q. 2023.
\newblock CTW: Confident Time-Warping for Time-Series Label-Noise Learning.
\newblock In \emph{IJCAI}, 4046--4054.

\bibitem[{Menon et~al.(2020)Menon, Rawat, Reddi, and Kumar}]{menon2020can}
Menon, A.~K.; Rawat, A.~S.; Reddi, S.~J.; and Kumar, S. 2020.
\newblock Can gradient clipping mitigate label noise?
\newblock In \emph{International Conference on Learning Representations}.

\bibitem[{Misawa, Suzuki, and Miura(2022)}]{misawa2022relationship}
Misawa, A.; Suzuki, A.; and Miura, H. 2022.
\newblock Relationship analysis between BCG features and blood pressure.
\newblock In \emph{2022 Joint 12th International Conference on Soft Computing and Intelligent Systems and 23rd International Symposium on Advanced Intelligent Systems (SCIS\&ISIS)}, 1--2. IEEE.

\bibitem[{Nagaraj et~al.(2024)Nagaraj, Gerych, Tonekaboni, Goldenberg, Ustun, and Hartvigsen}]{nagaraj2024learning}
Nagaraj, S.; Gerych, W.; Tonekaboni, S.; Goldenberg, A.; Ustun, B.; and Hartvigsen, T. 2024.
\newblock Learning from Time Series under Temporal Label Noise.
\newblock \emph{arXiv preprint arXiv:2402.04398}.

\bibitem[{Ogedegbe and Pickering(2010)}]{ogedegbe2010principles}
Ogedegbe, G.; and Pickering, T. 2010.
\newblock Principles and techniques of blood pressure measurement.
\newblock \emph{Cardiology clinics}, 28(4): 571--586.

\bibitem[{Pan et~al.(2024)Pan, Liang, Liang, Tang, Chen, and Zhu}]{pan2024robust}
Pan, J.; Liang, L.; Liang, Y.; Tang, Q.; Chen, Z.; and Zhu, J. 2024.
\newblock Robust modelling of arterial blood pressure reconstruction from photoplethysmography.
\newblock \emph{Scientific Reports}, 14(1): 1--13.

\bibitem[{Saeed et~al.(2011)Saeed, Villarroel, Reisner, Clifford, Lehman, Moody, Heldt, Kyaw, Moody, and Mark}]{saeed2011multiparameter}
Saeed, M.; Villarroel, M.; Reisner, A.~T.; Clifford, G.; Lehman, L.-W.; Moody, G.; Heldt, T.; Kyaw, T.~H.; Moody, B.; and Mark, R.~G. 2011.
\newblock Multiparameter Intelligent Monitoring in Intensive Care II: a public-access intensive care unit database.
\newblock \emph{Critical care medicine}, 39(5): 952--960.

\bibitem[{Sarkar and Etemad(2021)}]{sarkar2021cardiogan}
Sarkar, P.; and Etemad, A. 2021.
\newblock Cardiogan: Attentive generative adversarial network with dual discriminators for synthesis of ecg from ppg.
\newblock In \emph{Proceedings of the AAAI Conference on Artificial Intelligence}, volume~35, 488--496.

\bibitem[{Shu et~al.(2019)Shu, Xie, Yi, Zhao, Zhou, Xu, and Meng}]{shu2019meta}
Shu, J.; Xie, Q.; Yi, L.; Zhao, Q.; Zhou, S.; Xu, Z.; and Meng, D. 2019.
\newblock Meta-weight-net: Learning an explicit mapping for sample weighting.
\newblock \emph{Advances in neural information processing systems}, 32.

\bibitem[{Song et~al.(2022)Song, Kim, Park, Shin, and Lee}]{song2022learning}
Song, H.; Kim, M.; Park, D.; Shin, Y.; and Lee, J.-G. 2022.
\newblock Learning from noisy labels with deep neural networks: A survey.
\newblock \emph{IEEE transactions on neural networks and learning systems}, 34(11): 8135--8153.

\bibitem[{Wang et~al.(2023)Wang, Mohseni, Kilgore, and Najafizadeh}]{wang2023pulsedb}
Wang, W.; Mohseni, P.; Kilgore, K.~L.; and Najafizadeh, L. 2023.
\newblock PulseDB: A large, cleaned dataset based on MIMIC-III and VitalDB for benchmarking cuff-less blood pressure estimation methods.
\newblock \emph{Frontiers in Digital Health}, 4: 1090854.

\bibitem[{Wang et~al.(2018)Wang, Zhou, Noulas, Mascolo, Xie, and Chen}]{ijcai2018p497}
Wang, Y.; Zhou, X.; Noulas, A.; Mascolo, C.; Xie, X.; and Chen, E. 2018.
\newblock Predicting the Spatio-Temporal Evolution of Chronic Diseases in Population with Human Mobility Data.
\newblock In \emph{Proceedings of the Twenty-Seventh International Joint Conference on Artificial Intelligence, {IJCAI-18}}, 3578--3584. International Joint Conferences on Artificial Intelligence Organization.

\bibitem[{Wu et~al.(2021)Wu, Shu, Xie, Zhao, and Meng}]{wu2021learning}
Wu, Y.; Shu, J.; Xie, Q.; Zhao, Q.; and Meng, D. 2021.
\newblock Learning to purify noisy labels via meta soft label corrector.
\newblock In \emph{Proceedings of the AAAI Conference on Artificial Intelligence}, volume~35, 10388--10396.

\bibitem[{Xiao, Ding, and Hu(2022)}]{xiao2022time}
Xiao, R.; Ding, C.; and Hu, X. 2022.
\newblock Time Synchronization of Multimodal Physiological Signals through Alignment of Common Signal Types and Its Technical Considerations in Digital Health.
\newblock \emph{Journal of Imaging}, 8(5): 120.

\bibitem[{Xiao et~al.(2022)Xiao, Dong, Wang, Feng, Wu, Chen, and Zhao}]{xiao2022promix}
Xiao, R.; Dong, Y.; Wang, H.; Feng, L.; Wu, R.; Chen, G.; and Zhao, J. 2022.
\newblock Promix: Combating label noise via maximizing clean sample utility.
\newblock \emph{arXiv preprint arXiv:2207.10276}.

\bibitem[{Yuan et~al.(2024)Yuan, Wang, Li, Zhang, Hu, and Deen}]{yuan2024catransformer}
Yuan, X.; Wang, W.; Li, X.; Zhang, Y.; Hu, X.; and Deen, M.~J. 2024.
\newblock CATransformer: A Cycle-Aware Transformer for High-Fidelity ECG Generation From PPG.
\newblock \emph{IEEE Journal of Biomedical and Health Informatics}.

\bibitem[{Zeng et~al.(2023)Zeng, Chen, Zhang, and Xu}]{zeng2023transformers}
Zeng, A.; Chen, M.; Zhang, L.; and Xu, Q. 2023.
\newblock Are transformers effective for time series forecasting?
\newblock In \emph{Proceedings of the AAAI conference on artificial intelligence}, volume~37, 11121--11128.

\bibitem[{Zhang et~al.(2021)Zhang, Bengio, Hardt, Recht, and Vinyals}]{zhang2021understanding}
Zhang, C.; Bengio, S.; Hardt, M.; Recht, B.; and Vinyals, O. 2021.
\newblock Understanding deep learning (still) requires rethinking generalization.
\newblock \emph{Communications of the ACM}, 64(3): 107--115.

\bibitem[{Zhang(2017)}]{zhang2017mixup}
Zhang, H. 2017.
\newblock mixup: Beyond empirical risk minimization.
\newblock \emph{arXiv preprint arXiv:1710.09412}.

\bibitem[{Zhang et~al.(2024)Zhang, Zhu, Yang, Jin, Zhu, and Chen}]{zhang2024cross}
Zhang, Q.; Zhu, Y.; Yang, M.; Jin, G.; Zhu, Y.; and Chen, Q. 2024.
\newblock Cross-to-merge training with class balance strategy for learning with noisy labels.
\newblock \emph{Expert Systems with Applications}, 249: 123846.

\bibitem[{Zheltonozhskii et~al.(2022)Zheltonozhskii, Baskin, Mendelson, Bronstein, and Litany}]{zheltonozhskii2022contrast}
Zheltonozhskii, E.; Baskin, C.; Mendelson, A.; Bronstein, A.~M.; and Litany, O. 2022.
\newblock Contrast to divide: Self-supervised pre-training for learning with noisy labels.
\newblock In \emph{Proceedings of the IEEE/CVF Winter Conference on Applications of Computer Vision}, 1657--1667.

\bibitem[{Zheng, Awadallah, and Dumais(2021)}]{zheng2021meta}
Zheng, G.; Awadallah, A.~H.; and Dumais, S. 2021.
\newblock Meta label correction for noisy label learning.
\newblock In \emph{Proceedings of the AAAI conference on artificial intelligence}, volume~35, 11053--11061.

\end{thebibliography}
}

\clearpage
\appendix
\setcounter{secnumdepth}{1}

\section{Proof}  
\label{sec:proof}
\noindent \textbf{Proof of Eq.~\ref{eq:partial_theta_alpha}.}
To prove Eq.~\ref{eq:partial_theta_alpha}, we use the implicit dependence of $\theta$ on $\alpha$, where $\theta$, updated $k$ times per update of $\alpha$, can be expressed as a function of $\alpha$:
\begin{align}
\frac{\partial {\theta}^{{\tau}+1}}{\partial \alpha}  &= \frac{\partial}{\partial \alpha} ({\theta}^{{\tau}} - \eta \nabla_{\theta} {\mathcal{L}_{D'}}(\alpha,{\theta}^{{\tau}}(\alpha))) \\ 
&= \frac{\partial {\theta}^{{\tau}}}{\partial \alpha} - {\eta}\frac{\partial}{\partial \alpha}\nabla_{\theta} {\mathcal{L}_{D'}}(\alpha,{\theta}^{{\tau}}(\alpha))\\
& =  (I - {\eta}
  H_{{\theta},{\theta}}^{{\tau}})\frac{\partial {\theta}^{{\tau}}}{\partial \alpha} - {\eta} H_{{\theta},\alpha}^{{\tau}}\\
& \approx (1 - {\eta} )\frac{\partial {\theta}^{{\tau}}}{\partial \alpha} - {\eta} H_{{\theta},\alpha}^{{\tau}} \ .
\label{eq:partial_theta_alpha_app}
\end{align}

\noindent \textbf{Proof of Eq.~\ref{eq:partial_theta_alpha_exp_recursive}.}
Eq.~\ref{eq:partial_theta_alpha_app} establishes a recursive relation for $\frac{\partial \theta^{\tau+1}}{\partial \alpha}$, expressing it in terms of $\frac{\partial \theta^\tau}{\partial \alpha}$ and higher-order gradients. Opening this recurrence, we derive:
\begin{align}
\frac{\partial {\theta}^{\tau+1}}{\partial \alpha}  
&=(1 - {\eta} )^k\frac{\partial {\theta}^{\tau-k+1}}{\partial \alpha} - {\eta}\sum_{j=0}^{k-1}(1 - {\eta} )^jH_{{\theta},\alpha}^{\tau-j}  \\
&\approx - {\eta}\sum_{j=0}^{k-1}(1 - {\eta} )^jH_{{\theta},\alpha}^{\tau-j}\\
&= -  {\eta} H_{{\theta},\alpha}^{\tau} - {\eta}\sum_{j=1}^{k-1}(1 - {\eta} )^jH_{{\theta},\alpha}^{\tau-j}.
\end{align}

\noindent \textbf{Proof of Eq.~\ref{eq:k_step_meta_gradient}.}
In the following proof, $\frac{\partial {\theta}^{\tau+1}}{\partial \alpha} $ is substituted into the meta-gradient computation. The past $k-1$ gradients $g_{{\theta}^{\tau-j+1}}$ ($1\leq j\leq k-1$) are approximated by the current gradient $g_{{\theta}^{\tau+1}}$, avoiding storage of historical gradients.
\begin{align}
     \frac{\partial {\mathcal{L}_{D}}{({\theta}^{\tau+1})}}{\partial \alpha} 
&= g_{{\theta}^{\tau+1}}  \frac{\partial {\theta}^{\tau+1}}{\partial \alpha} \\
& = -  {\eta}g_{{\theta}^{\tau+1}} H_{{\theta},\alpha}^{\tau} + {\eta} \sum_{j=1}^{k-1} \gamma^{\tau-j}  (-  g_{{\theta}^{\tau-j+1}} H_{{\theta},\alpha}^{\tau-j}) \\
& \approx  -  {\eta}g_{{\theta}^{\tau+1}} H_{{\theta},\alpha}^{\tau} +   (1-{\eta})\frac{\partial {\mathcal{L}_{D}}{({\theta}^{\tau})}}{\partial \alpha} \  , 
 \end{align}
\noindent where $\gamma^{\tau-j} = g_{{\theta}^{\tau+1}}(1 - {\eta} )^{j} \frac{g_{{\theta}^{\tau-j+1}}^{\top}}{\|g_{{\theta}^{\tau-j+1}}\|^2}  $.

\section{Supplementary Experimental Material}
\label{app-Experimental}
\subsection{Baseline Details}
Here’s a detailed introduction to the baselines:

Basic Backbone Models:
\begin{itemize}
\item The basic baselines are models for waveform transformation that does not handle corrupted data, including Swin-Transformer~\cite{liu2021swin}, InceptionTime~\cite{ismail2020inceptiontime}, and ResNet~\cite{he2016deep}. 
\end{itemize} 

Sample Selection Methods:
\begin{itemize}
    \item  MW-Net~\cite{shu2019meta}: A meta-learning approach that adaptively learns loss weights to emphasize clean samples consistent with meta-knowledge.
    \item  Co-teaching~\cite{han2018co}: Trains two networks that mutually select likely clean samples in each mini-batch.
\end{itemize} 

Label Correction Methods:
\begin{itemize}
    \item U-correction~\cite{arazo2019unsupervised}: Estimates label corruption probabilities, uses bootstrapping loss and adapts Mixup augmentation.
    \item DivideMix~\cite{li2020dividemix}: Uses Gaussian Mixture Models to classify clean and noisy samples and uses two networks iteratively for label co-refinement and label co-guessing on labeled and unlabeled samples.
    \item MSLC~\cite{wu2021learning}: Learns a weight network based on meta-learning to generate soft labels by combining noisy labels with the backbone’s historical predictions.    
    \item MLC~\cite{zheng2021meta}: Directly generates corrected labels for noisy samples using a meta-learning-based label correction network.
    \item C2MT~\cite{zhang2024cross}: Based on DivideMix’s cross-training, C2MT periodically merges the two networks into a single model via federated parameter averaging.
\end{itemize}

Both sample selection and label correction methods are model-agnostic, using InceptionTime as the backbone for consistency in our experiments, though others can be used.

\subsection{Parameter Sensitivity} 
\label{sec:Parameter-Sensitivity} 
\begin{figure}[!hb]
\centering 
    \subfloat[\label{sensi-kstep_vital}VitalDB.]{\includegraphics[width=.48\columnwidth]{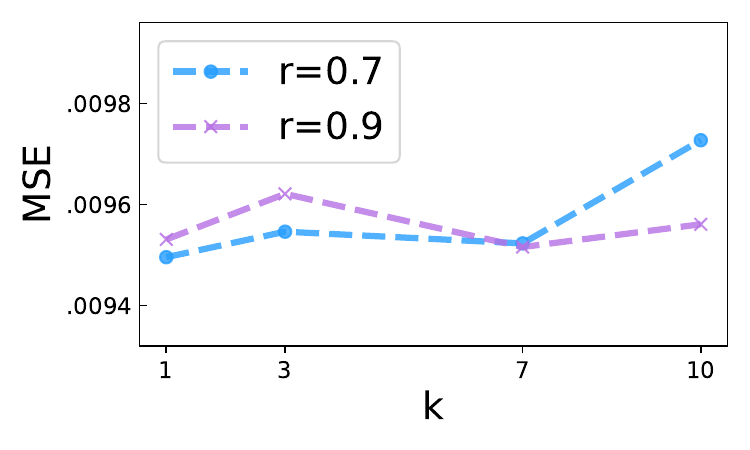}} 
    \subfloat[\label{sensi-kstep_oml}OML.]{\includegraphics[width=.48\columnwidth]{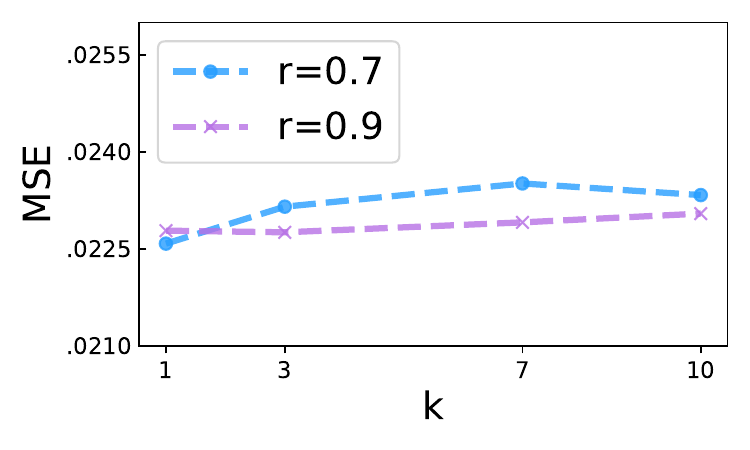}} 
\caption{Parameter analysis on $k$. }
\label{sensi-k}
\end{figure}

We evaluate the sensitivity of  $k$ on VitalDB and OML in Fig~\ref{sensi-k}. Results show that  $k$  performs well when under 10. However, as Fig.~\ref{sensi-kstep_vital} shows, for  $r = 0.7$, smaller  $k$  yield better approximations, whereas with a higher corruption rate of  $r = 0.9$, longer steps improve performance, supporting the effectiveness of  k-step meta-gradient approximation.

We further analyzed the effect of metaset size $M$, a small practical set of manually aligned pairs (1-5\% of the training data). As shown in Table~\ref{tab:sensi-M}, performance on the VitalDB dataset plateaus beyond $M=500$, indicating \textit{SyncNet}’s sufficient correction.

\begin{table}[ht!]
    \centering
     { \small
        \begin{tabular}{lccccc}
            \toprule
           \textbf{$M$} & 100 & 200 & 500 & 1000 & 2000  \\
           \midrule
            \textbf{MSE} & 0.01064 & 0.01007 & 0.00945 & 0.00944 & 0.00944 \\
            \bottomrule
        \end{tabular}
    }
    \caption{Parameter analysis on $M$.  }  
    \label{tab:sensi-M}
\end{table}

\subsection{Time-shift Tolerance Analysis}
\label{sec:Time-shift-Tolerance-Analysis} 

As shown in Table~\ref{tab:robust-change_rate},~\ref{tab:robust-change_shift}, and~\ref{tab:robust-change_samplerate}, we comprehensively evaluate the robustness of our method under varying corruption ratios ($r$), time-shift magnitudes ($S$), and training sample sizes ($N_r$) on the VitalDB and OML datasets, comparing it against state-of-the-art baselines  the sample selection (Co-teaching) and label correction (MLC) groups. Across all settings, our method consistently achieves the lowest MSE mean and standard deviation, demonstrating superior robustness and stability. Specifically, as $r$ rises  0.3 to 0.9, $S$ increases  5 to 40, or $N_r$ drops  0.9 to 0.3, our method maintains optimal performance, whereas Co-teaching and MLC exhibit noticeable performance degradation due to their dependence on clean sample subsets or limited label correction capability. Interestingly, under fixed $r = 0.7$ and $N_r = 1.0$, Co-teaching occasionally benefits  larger $S$, possibly because larger shifts make it easier to identify and eliminate misaligned samples. However, its performance drops sharply when $r$ or $N_r$ reaches 0.9 or 0.3, owing to insufficient clean supervision. In contrast, our method leverages the full dataset via effective time-shift correction by \textit{SyncNet}, consistently ensuring robust performance even under severe misalignment and data scarcity.
 
\begin{table}[ht!]
\centering
    {\small
    \begin{tabular}{llccc}
    \toprule
    {Dataset} & $r$ & {Co-teaching} & {MLC} & {Ours} \\
    \midrule
    \multirow{6}{*}{{VitalDB}} 
    & $0.3$ & 0.0086 & 0.0121 & \textbf{0.0095} \\
    &$ 0.5$ & 0.0092 & 0.0123 & \textbf{0.0093} \\
    & $0.7$ & 0.0103 & 0.0133 & \textbf{0.0095} \\
    & $0.9$ & 0.0159 & 0.0140 & \textbf{0.0095} \\
    \cmidrule(lr){2-5}
    & Avg & 0.0110 & 0.0129 & \textbf{0.0095} \\
    & Std & 0.0033 & 0.0009 & \textbf{0.0001} \\
    \midrule
    \multirow{6}{*}{{OML}} 
    & $0.3$ & 0.0244 & 0.0309 & \textbf{0.0241} \\
    & $0.5$ & 0.0243 & 0.0314 & \textbf{0.0221} \\
    & $0.7$ & 0.0252 & 0.0324 & \textbf{0.0226} \\
    & $0.9$ & 0.0297 & 0.0337 & \textbf{0.0228} \\
    \cmidrule(lr){2-5}
    & Avg & 0.0259 & 0.0321 & \textbf{0.0229} \\
    & Std & 0.0026 & 0.0012 & \textbf{0.0009} \\
    \bottomrule
    \end{tabular}
 }
 \caption{Test MSE under rising $r$ \normalfont {$(S=20,N_r=1.0)$}.}
    \label{tab:robust-change_rate}
\end{table}

\begin{table}[ht!]
\centering
 
    {\small
        \begin{tabular}{llccc}
        \toprule
        {Dataset} & $S$ & {Co-teaching} & {MLC} & {Ours} \\
        \midrule
        \multirow{8}{*}{{VitalDB}} 
        & $5$  & 0.0120 & 0.0117 & \textbf{0.0096} \\
        & $10$ & 0.0114 & 0.0122 & \textbf{0.0093} \\
        & $15$ & 0.0109 & 0.0126 & \textbf{0.0095} \\
        & $20$ & 0.0103 & 0.0133 & \textbf{0.0095} \\
        & $30$ & 0.0101 & 0.0144 & \textbf{0.0097} \\
        & $40$ & \textbf{0.0095} & 0.0141 & {0.0098} \\
        \cmidrule(lr){2-5}
        & Avg & 0.0107 & 0.0131 & \textbf{0.0096} \\
        & Std & 0.0009 & 0.0011 & \textbf{0.0002} \\
        \midrule
        \multirow{8}{*}{{OML}} 
        & $5$  & 0.0255 & 0.0306 & \textbf{0.0219} \\
        & $10$ & 0.0271 & 0.0310 & \textbf{0.0222} \\
        & $15$ & 0.0261 & 0.0315 & \textbf{0.0225} \\
        & $20$ & 0.0252 & 0.0324 & \textbf{0.0226} \\
        & $30$ & 0.0241 & 0.0345 & \textbf{0.0232} \\
        & $40$ & 0.0250 & 0.0334 & \textbf{0.0247} \\
        \cmidrule(lr){2-5}
        & Avg & 0.0255 & 0.0322 & \textbf{0.0228} \\
        & Std & 0.0010 & 0.0015 & \textbf{0.0010} \\

        \bottomrule
        \end{tabular} 
    }
    \caption{Test MSE under rising $S$ \normalfont {$(r =0.7,N_r=1.0)$}.} 
    \label{tab:robust-change_shift}
\end{table}

\begin{table}[ht!]
\centering
 \centering
     
    {\small
        \begin{tabular}{llccc}
        \toprule
        {Dataset} & $N_r$ & {Co-teaching} & {MLC} & {Ours} \\
        \midrule
        \multirow{6}{*}{{VitalDB}} 
        & $0.9$ & 0.0104 & 0.0117 & \textbf{0.0095} \\
        & $0.7$ & 0.0109 & 0.0124 & \textbf{0.0095} \\
        & $0.5 $& 0.0122 & 0.0115 & \textbf{0.0097} \\
        & $0.3$ & 0.0141 & 0.0121 & \textbf{0.0101} \\
        \cmidrule(lr){2-5}
        & Avg & 0.0119 & 0.0119 & \textbf{0.0097} \\
        & Std & 0.0016 & 0.0004 & \textbf{0.0003} \\
        \midrule
        \multirow{6}{*}{{OML}} 
        & $0.9 $& 0.0262 & 0.0327 & \textbf{0.0233} \\
        & $0.7$ & 0.0262 & 0.0339 & \textbf{0.0241} \\
        & $0.5$ & 0.0276 & 0.0350 & \textbf{0.0238} \\
        & $0.3$ & 0.0287 & 0.0356 & \textbf{0.0247} \\
        \cmidrule(lr){2-5}
        & Avg & 0.0272 & 0.0343 & \textbf{0.0240} \\
        & Std & 0.0012 & 0.0013 & \textbf{0.0006} \\
        \bottomrule
        \end{tabular} 
    }
    \caption{Test MSE under decreasing $N_r$ \normalfont{ ($S=20,r=0.7$)}.}
    \label{tab:robust-change_samplerate}
\end{table}

\subsection{Efficiency Analysis} 
Our method adds moderate overhead during training but incurs no additional cost during inference. Specifically, training involves two extra operations: (1) \textit{SyncNet} forward on  $D'$, and (2) \textit{TransNet} updates on $D$ combined with \textit{SyncNet}’s meta-updates every $k$ steps. These factors raise training time by 2-3×. In terms of memory, storing one extra batch of D and the \textit{SyncNet} parameters incurs limited overhead, and the space-efficient implementation of Eq.~\ref{eq:k_step_meta_gradient} requires storing only the latest meta-gradient. At inference, only \textit{TransNet} is used, adding no extra time or memory cost.

\subsection{Implementation} 
\label{sec:implementation}
Both our method and the baselines for handling corrupted samples utilize InceptionTime as the backbone network. In \textit{SyncNet}, we stack convolutional layers with a kernel size of 15 and LeakyReLU activation across the two convolutional blocks, followed by a three-layer MLP for time-shift estimation. 
We use a batch size of 128, with initial learning rates of 1.5e-3 for \textit{TransNet} and 5e-5 for \textit{SyncNet}. The Adam optimizer is used for parameter updates, with a weight decay of 5e-4 for \textit{TransNet}. A cosine annealing schedule is applied to adjust the learning rate. We use the first 10 epochs to warmup \textit{TransNet}. For models that require the metaset, we randomly select 500, 500, and 300 signal segments  the VitalDB, MIMIC, and OML datasets, respectively.
Experiments are conducted on an NVIDIA A40 GPU with Ubuntu 20.04.6 LTS, using Python 3.10.0, PyTorch 1.13.0.

\subsection{Future Extensions} 
Our meta-learning bi-level optimization framework provides a basis for applications beyond time-series misalignment. Its modular, model-agnostic design supports extensions to other forms of label corruption, such as artifacts, missing points, or multimodal misalignment in diverse domains like audiovisual synchronization and biomedical signal processing.

\end{document}